\documentclass{article} % For LaTeX2e
\usepackage{iclr2026_conference,times}
\iclrfinalcopy

\usepackage{amsmath,amsfonts,bm,eqnarray}
\usepackage{cancel}
\usepackage{amsthm}
\usepackage{tikz}
\usetikzlibrary{tikzmark}

\newtheorem{innercustomgeneric}{\customgenericname}
\providecommand{\customgenericname}{}
\newcommand{\newcustomtheorem}[2]{%
  \newenvironment{#1}[1]
  {%
   \renewcommand\customgenericname{#2}%
   \renewcommand\theinnercustomgeneric{##1}%
   \innercustomgeneric
  }
  {\endinnercustomgeneric}
}

\newcustomtheorem{customThm}{Theorem}
\newcustomtheorem{customLemma}{Lemma}
\newcustomtheorem{customCor}{Corollary}
\newcustomtheorem{customProposition}{Proposition}

\def\Tabref#1{Table~\ref{#1}}

\def\Figref#1{Fig.~\ref{#1}}

\def\appref#1{Appendix~\ref{#1}}
\def\Secref#1{Sec.~\ref{#1}}
\def\eqref#1{equation~\ref{#1}}
\def\Eqref#1{Eqn.~\ref{#1}}

\def\Algref#1{Algorithm~\ref{#1}}

\def\1{\bm{1}}

\def\rvx{{\mathbf{x}}}

\def\vtheta{{\bm{\theta}}}

\def\va{{\bm{a}}}

\def\vc{{\bm{c}}}

\def\vq{{\bm{q}}}

\def\vs{{\bm{s}}}

\def\vx{{\bm{x}}}
\def\vy{{\bm{y}}}
\def\vz{{\bm{z}}}

\def\mA{{\bm{A}}}

\def\mC{{\bm{C}}}

\def\mS{{\bm{S}}}

\def\mX{{\bm{X}}}

\DeclareMathAlphabet{\mathsfit}{\encodingdefault}{\sfdefault}{m}{sl}
\SetMathAlphabet{\mathsfit}{bold}{\encodingdefault}{\sfdefault}{bx}{n}

\renewcommand{\hat}{\widehat}
\renewcommand{\frac}{\tfrac}

\usepackage{microtype}
\usepackage{graphicx}
\usepackage{subfigure}
\usepackage{wrapfig,lipsum,booktabs}       % professional-quality
\usepackage{amsfonts}       % blackboard math symbols
\usepackage{nicefrac}       % compact symbols for 1/2, etc.
\usepackage[dvipsnames]{xcolor}  
\usepackage{colortbl}
\usepackage{amssymb}
\usepackage{pifont}
\usepackage{rotating}
\usepackage{makecell}

\usepackage{bbm}

\usepackage{caption}
\usepackage{subcaption}
\usepackage{amsmath}
\usepackage{xspace}
\usepackage{multirow}

\usepackage[titletoc]{appendix}

\usepackage{pdflscape}

\usepackage{tabularray}

\usepackage{soul}
\usepackage[dvipsnames]{xcolor} % For more color options
\usepackage{tcolorbox}
\tcbuselibrary{skins,listings,breakable} % Needed for enhanced styling

\usepackage[colorlinks=true,citecolor=brown,urlcolor=gray,linkcolor=BlueViolet]{hyperref}

\newtoggle{comments}
\toggletrue{comments}
\iftoggle{comments}{
    \newcommand{\semih}[1]{\textcolor{blue}{(Semih: #1)}}
}{
    \newcommand{\semih}[1]{}
}

\iftoggle{comments}{
    \newcommand{\yingbo}[1]{\textcolor{cyan}{(YB: #1)}}
}{
    \newcommand{\yingbo}[1]{}
}
\iftoggle{comments}{
    \newcommand{\zeyu}[1]{\textcolor{green}{(ZY: #1)}}
}{
    \newcommand{\zeyu}[1]{}
}
\allowdisplaybreaks

\usepackage[compact]{titlesec}
\usepackage{sidecap}

\usepackage{enumitem}
\usepackage{arydshln} 
\usepackage[utf8]{inputenc} % allow utf-8 input
\usepackage[T1]{fontenc}    % use 8-bit T1 fonts

\newcommand{\StartMenu}{\raisebox{-0.15cm}{\includegraphics[scale=0.007]{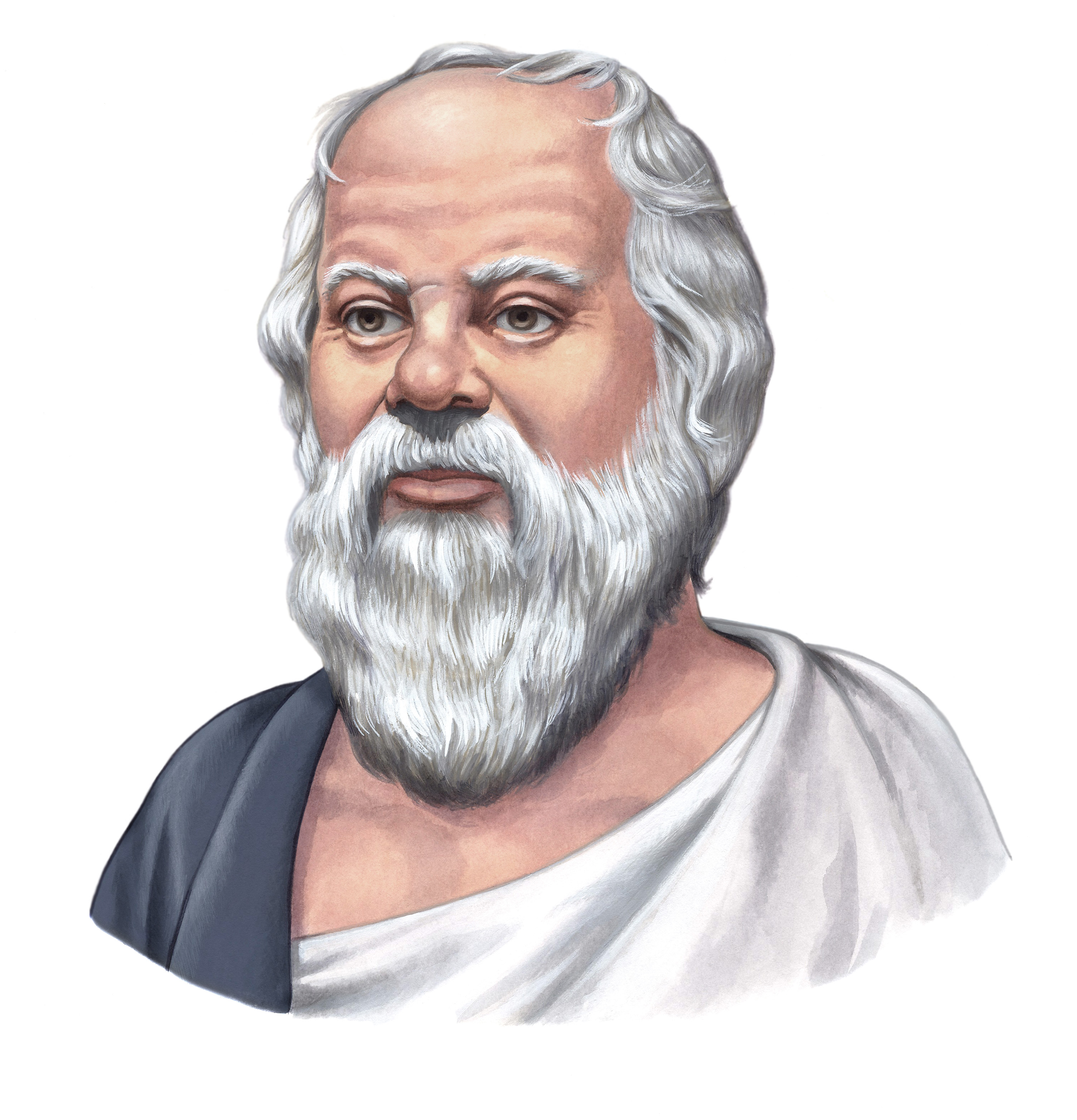}}}%
\newcommand{\DRAW}[1]{%
   \StartMenu
   \foreach \x in {#1} {%
   \texttt{\x}%    
   }%
}%

\newcommand{\ours}{\texttt{SSR}\xspace}

\newcommand{\gptnano}{\texttt{GPT-4.1-nano}\xspace}
\newcommand{\gptfive}{\texttt{GPT-5-mini}\xspace}
\newcommand{\geminiflash}{\texttt{Gemini-2.5-Flash}\xspace}
\newcommand{\geminilite}{\texttt{Gemini-2.5-Flash-Lite}\xspace}

\usepackage{siunitx}
\newlength\savewidth

\newcolumntype{C}{>{\centering\let\newline\\\arraybackslash\hspace{0pt}}m{2cm}}
\definecolor{lightergray}{HTML}{e5e5e5}
\sethlcolor{lightergray}

\title{\vspace{-0.12cm}\DRAW{\ours}: Socratic Self-Refine for Large Language Model Reasoning}

\newcommand{\SF}[0]{\textbf{\textsuperscript{1}}}
\newcommand{\RU}[0]{\textbf{\textsuperscript{2}}}
\newcommand{\UT}[0]{\textbf{\textsuperscript{3}}}
\newcommand{\authorsep}[0]{\ \ }

\author{\leavevmode\unskip
\textbf{Haizhou Shi}\thanks{
 Work done during an internship at Salesforce AI Research.
 $^{\textbf{\textdagger}}$Correspondence to: 
  Haizhou Shi <haizhou.shi@rutgers.edu>, Semih Yavuz <syavuz@salesforce.com>.
} \space$^{\text{\textdagger}}$\SF\RU\authorsep
\textbf{Ye Liu}\SF\authorsep
\textbf{Bo Pang}\SF\authorsep
\textbf{Zeyu Leo Liu}$^{*}$\SF\UT\authorsep
\textbf{Hao Wang}\RU\authorsep
\\
\textbf{Silvio Savarese}\SF\authorsep
\textbf{Caiming Xiong}\SF\authorsep
\textbf{Yingbo Zhou}\SF\authorsep
\textbf{Semih Yavuz}$^{\textbf{\textdagger}}$\SF
\\
\SF Salesforce AI Research \authorsep
\RU Rutgers University \authorsep
\UT The University of Texas at Austin 
}

\begin{document}

\maketitle

\begin{abstract}
Large Language Models~(LLMs) have demonstrated remarkable reasoning abilities, yet existing test-time frameworks often rely on coarse self-verification and self-correction, limiting their effectiveness on complex tasks. 
In this paper, we propose \textbf{S}ocratic \textbf{S}elf-\textbf{R}efine (\textbf{\ours}), a novel framework for fine-grained evaluation and precise refinement of LLM reasoning. Our proposed \ours decomposes model responses into verifiable (sub-question, sub-answer) pairs, enabling step-level confidence estimation through controlled re-solving and self-consistency checks. By pinpointing unreliable steps and iteratively refining them, \ours produces more accurate and interpretable reasoning chains. Empirical results across five reasoning benchmarks and three LLMs show that \ours consistently outperforms state-of-the-art iterative self-refinement baselines. Beyond performance gains, \ours provides a principled black-box approach for evaluating and understanding the internal reasoning processes of LLMs.
{Code is available at \url{https://github.com/SalesforceAIResearch/socratic-self-refine-reasoning}.}
\end{abstract}

\begin{figure}[h] 
\centering 
\includegraphics[width=1\textwidth]{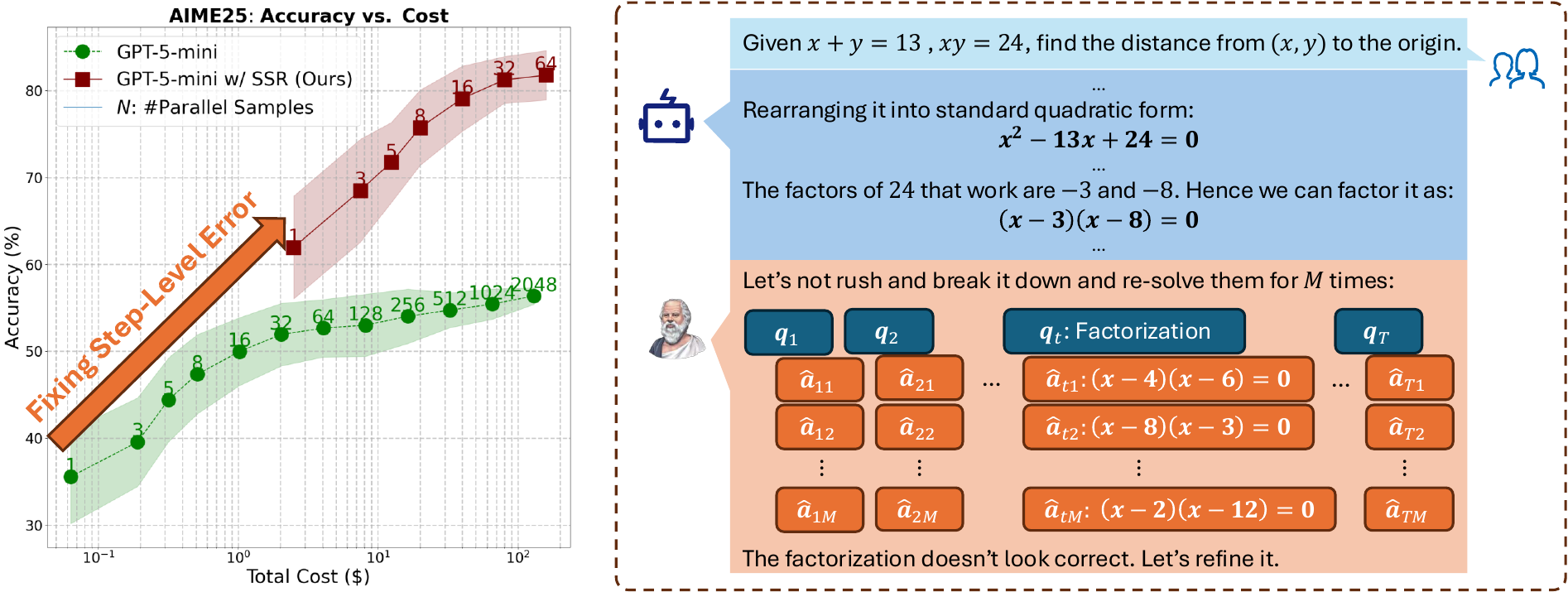}
\caption{
    \textbf{Test-Time Parallel Scaling Performance (Left)} and \textbf{Conceptual Overview (Right)} of our proposed \textbf{S}ocratic \textbf{S}elf-\textbf{R}efine \textbf{(\ours)}.
    By decomposing responses into Socratic steps, re-evaluating intermediate results through self-consistency, and refining specific step-level errors, \ours achieves substantially higher initial accuracy \textbf{($\sim$67.57\% relative improvement)} and continues to scale effectively even when standard Chain-of-Thought (CoT) begins to saturate. Notably, this performance advantage holds under comparable computational cost. Experiments are conducted with \gptfive in low-reasoning, low-verbosity mode.
} 
\label{fig:figure1} 
\end{figure}

\section{Introduction}
\label{sec:intro}
Large Language Models~(LLMs) have rapidly advanced the frontier of machine reasoning, demonstrating impressive performance across domains ranging from mathematical problem solving to complex logical inference~\citep{wei2022emergent,wang2022self,chung2024scaling,guo2025deepseek,ke2025survey}. Central to these capabilities is the paradigm of reasoning with explicit intermediate steps, often instantiated through chain-of-thought (CoT) prompting~\citep{wei2022chain}. By externalizing reasoning traces, CoT enables models to articulate their latent decision-making process, offering both interpretability and opportunities for iterative improvement~\citep{madaan2023self}. Despite these strengths, the reasoning traces generated by LLMs remain prone to cascading errors: a single flawed step can propagate downstream, leading to incorrect or incoherent final answers~\citep{wu2025more,you2025probabilistic}. This vulnerability raises pressing questions about how to reliably evaluate, refine, and searching for better multi-step reasoning at test time. 

Existing frameworks have sought to address these challenges largely fall into two paradigms: sample selection with self-verification and self-refinement. Sample selection with self-verification, aims to assess response reliability by assigning confidence scores to completed reasoning traces either by LLM-as-a-Judge~\citep{gu2024survey}, or a specific ranking model~\citep{snell2024scaling}, and then through multiple sampling and selection improves the final answer reliability~\citep{zheng2023judging,chen2025sets}. While these approaches can identify low-quality outputs, they often operate at a coarse granularity, overlooking subtle step-level errors embedded within long derivations~\citep{fang2025graph}. Self-refinement methods, by contrast, encourage LLMs to iteratively critique and revise their own responses~\citep{madaan2023self,zhang2024accessing,bi2024forest}. Although such frameworks have yielded measurable gains, their reliance on holistic self-feedback frequently limits their ability to pinpoint and correct specific erroneous steps. As a result, both paradigms struggle to provide robust and interpretable error correction in complex reasoning tasks.

In this paper, we propose \textbf{S}ocratic \textbf{S}elf-\textbf{R}efine (\textbf{\ours}), a novel framework designed to overcome these limitations by introducing fine-grained, step-level evaluation and targeted refinement of LLM reasoning. \ours reformulates the reasoning process into a sequence of verifiable (sub-question, sub-answer) pairs, which we refer to as Socratic steps. This decomposition enables precise confidence estimation through controlled re-solving and self-consistency checks at the step level. 
Unreliable steps are selectively refined, allowing the model to fix errors without depending on vague feedback.
By iteratively applying this process, \ours improves both the accuracy and interpretability of LLM reasoning, offering a principled black-box approach to evaluating and refining model behavior. 

Empirical results across 5 reasoning tasks (3 mathematical and 2 logical) and multiple state-of-the-art LLMs demonstrate that \ours consistently outperforms baseline self-refinement methods. Beyond raw accuracy gains, our analysis shows that \ours yields more reliable refinement trajectories, particularly when combined with plan-level adjustments or adaptive gating mechanisms. These findings highlight the importance of explicit step-level verification in building trustworthy LLM reasoning systems. More broadly, \ours represents a step toward interpretable and controllable test-time reasoning, bridging the gap between coarse-grained judgment and fine-grained error correction. To summarize, our contributions are: 
\begin{itemize}[nosep,leftmargin=24pt]
    \item We propose a novel framework, Socratic Self-Refine~(\ours), that allows more fine-grained confidence estimation and precise error control over decomposed reasoning steps.
    By formulating reasoning as a sequence of (sub-question, sub-answer) pairs, \ours overcomes the limitations of existing holistic self-refinement methods.
    \item We empirically validate \ours on 5 reasoning tasks using two state-of-the-art models, demonstrating that it consistently outperforms existing self-refine-based baselines.
    \item Our \ours introduces a mechanism for eliciting the model's step-level confidence, by having the LLM re-solve each sub-question multiple times with explicit context control. Leveraging self-consistency as a reliable confidence estimate for each step, \ours provides a pioneering effort in evaluating and interpreting the internal reasoning processes of LLMs.
\end{itemize}

\section{Related Work}
\label{sec:related}

\textbf{Self-Evaluation and Refinement of LLMs.}\quad
Recent work has introduced both \emph{intrinsic} and \emph{generative} approaches for LLM self-evaluation. On the intrinsic side, uncertainty-based methods estimate correctness either through consistency, by comparing multiple independently generated outputs~\citep{kuhn2023semantic, manakul2023selfcheckgpt}, or through statistics derived from the model’s output distribution~\citep{kang2025scalable,fu2025deep,zhang2025token}. On the generative side, the \emph{LLM-as-a-Judge} paradigm directly prompts models to evaluate responses, often achieving strong alignment with human preferences and supporting test-time strategies like abstaining from low-quality responses or selecting among candidates~\citep{zheng2023judging,gu2024survey,zhou2025evaluating,ren2023self,chen2025sets,huang2025efficient,zhong2025solve,zhou2025variation}. While limitations such as positional bias~\citep{zheng2023large,shi2024judging} and a preference for longer responses~\citep{hu2024explaining} do exist, both uncertainty-based and judge-based methods remain effective and have proven valuable for evaluating LLM outputs.
Building on these evaluation techniques, a growing body of work extends beyond verification to self-refinement, where LLMs not only diagnose weaknesses in their outputs but also iteratively improve them~\citep{madaan2023self}. Early efforts explored direct self-correction based on feedback, while subsequent methods introduced structured search~\citep{zhang2024accessing}, parallel sampling to enrich candidate diversity~\citep{bi2024forest,chen2025sets}, and reformulation strategies that generate improved sub-questions by incorporating contextual preconditions~\citep{teng2025atom}. More recent work trains generative verifiers to guide the refinement process~\citep{zhong2025solve}. Collectively, these approaches demonstrate that refinement transforms passive evaluation into an active mechanism for improving reliability, making it a key step toward controllable and trustworthy reasoning in LLMs.

\textbf{Process Evaluation of LLMs.}\quad
Verifying only the final outcome of an LLM is insufficient; ensuring reliability requires mechanisms that also evaluate the reasoning process itself. Beyond using human annotations to train process reward models~\citep{lightman2023let,skywork2024prm,zhang2025lessons}, the rapid advancement of model capabilities has motivated a growing set of test-time methods for step-level verification. These approaches typically decompose the reasoning trace and assess the correctness of each step to localize errors more accurately~\citep{ling2023deductive,zhao2025genprm,mukherjee2025premise,fang2025graph}.
Compared to existing work of process evaluation, our \ours framework adopts a Socratic formulation of reasoning, representing the process as a sequence of question-answer pairs~(details in \Secref{sec:socratic-self-refine}). This structure makes the steps straightforward to re-execute and enables reliable confidence estimation. Crucially, \ours goes beyond verification by producing informative signals that directly support subsequent refinement.

\section{Socratic Self-Refine (\ours)}
\label{sec:socratic-self-refine}

This section introduces our Socratic Self-Refine~(\ours). 
\Secref{sec:method-preliminary} introduces the fundamental assumption that natural-language reasoning can be described as a Socratic process.
\Secref{sec:method-ssr} presents the core of \ours, including the decomposition into Socratic steps, their verification, and reasoning refinement guided by Socratic confidence scores.
Finally, \Secref{method:ssr-deployment} discusses two techniques for practical deployment of \ours: plan refinement and adaptive iteration refinement.
\textbf{For details of the prompt templates introduced in this section, please refer to \appref{app:prompts}.}

\textbf{Notation.}\quad
In this paper, scalars are denoted by lowercase letters ($x$), vectors (or token/word sequences) by bold lowercase letters ($\vx$), random vectors by boldface lowercase letters ($\rvx$), and matrices (or sets of tokens, words, or phrases) by bold uppercase letters ($\mX$).
We denote by $[m]={1,2,\ldots,m}$ the set of consecutive integers from $1$ to $m$.
For consistency, $K$ denotes the total number of refinement iterations, \textbf{while $(k)$ indicates the current iteration; when unambiguous, we omit $(k)$ to reduce clutter.}
Finally, $N$ is the number of parallel runs used for test-time scaling.

\subsection{LLM Reasoning as Socratic Process}
\label{sec:method-preliminary}
\textbf{Preliminary of LLM Reasoning.}\quad
For problems with short-form ground-truth answers, LLM reasoning can be modeled as marginalization over intermediate natural language reasoning traces $\vz$ (a sequence of tokens/words) to produce the final answer $\vy$~\citep{chen2024language}:
\begin{align}
    \label{eq:reasoning-as-integration}
    \pi_\vtheta(\vy \mid \vx) &= \int \pi_\vtheta(\vy\mid\vz,\vx) \pi_\vtheta(\vz\mid\vx) d\vz 
    % \approx \frac{1}{N}\sum_{n=1}^N \pi_\vtheta(\vy\mid\vz_n,\vx), \vz_n\sim \pi_\vtheta(\vz\mid\vx).
\end{align}
Chain-of-Thought (CoT) reasoning~\citep{wei2022chain} approximates this integral with a single sample: the model first generates a reasoning trace $\vz \sim \pi_\vtheta(\cdot \mid \vx)$ and then derives the final answer $\vy \sim \pi_\vtheta(\cdot \mid \vz,\vx)$. Empirically, allocating more computation to approximate \Eqref{eq:reasoning-as-integration} improves performance. A common strategy is Majority Voting (Maj@N), which averages over multiple sampled reasoning traces~\citep{wang2022self}:
\begin{align}
    \pi_\vtheta(\vy \mid \vx) 
    % &= \int \pi_\vtheta(\vy\mid\vz,\vx) \pi_\vtheta(\vz\mid\vx) d\vz 
    &\approx \frac{1}{N}\sum\nolimits_{n=1}^N \pi_\vtheta(\vy\mid\vz_n,\vx), \quad \vz_n\sim \pi_\vtheta(\vz\mid\vx).
\end{align}

\begin{figure}[t] 
\centering 
\includegraphics[width=1\textwidth]{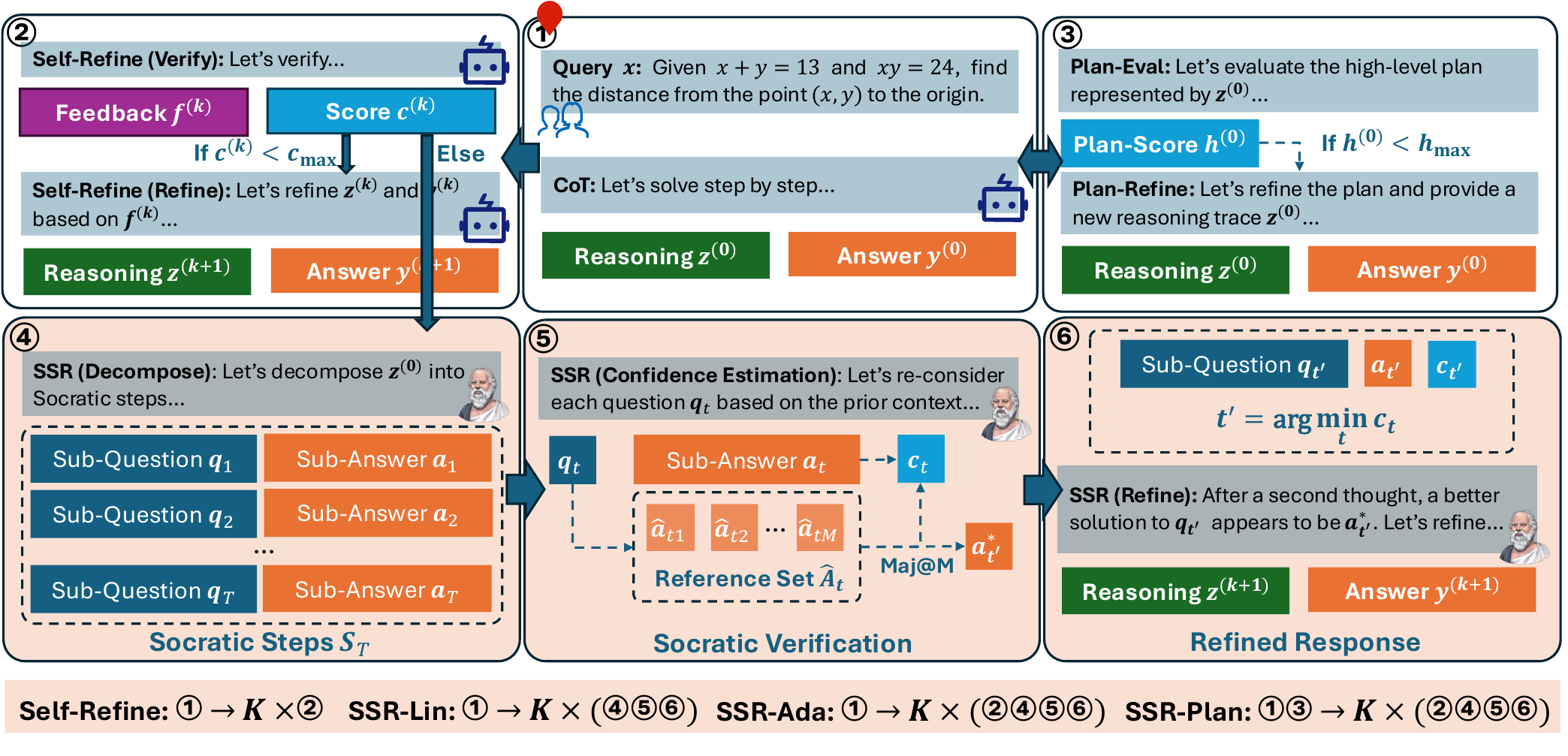}
\vspace{-2em}
\caption{
    Overview of \textbf{S}ocratic \textbf{S}elf-\textbf{R}efine~(\textbf{\ours}). 
    \textbf{Block \ding{172}:} Chain-of-Thought (CoT) reasoning, serves as the starting point for the iterative refinement methods; 
    \textbf{Block \ding{173}:} Simple Self-Refine, generates feedback and then refines the original response based on the feedback;
    \textbf{Block \ding{174}:} Plan refinement, summarizes the high-level plan of a reasoning trace, and refines the plan and the trace if necessary;
    \textbf{Block \ding{175}-\ding{177}:} Three building blocks of our \ours, includes Socratic decomposition, Socratic verification, and Socratic refinement. 
    \textbf{\ours-Lin}: Linear \ours, faithfully applies three blocks (\ding{175}-\ding{177}) for $K$ iterations;
    \textbf{\ours-Ada}: Adaptive \ours, only carries out Socratic blocks (\ding{175}-\ding{177}) when the normal Self-Refine cannot identify any mistakes ($\vc=\vc_{\max}$);
    \textbf{\ours-Plan}: Adaptive \ours with Plan Refinement, adds an additional plan refinement round (\ding{174}) before the full iterative refinement algorithm (\ding{175}-\ding{177}).
} 
\vspace{0.5em}
\label{fig:overview} 
\end{figure}

\textbf{Reasoning as Socratic Process.}\quad
In this paper, we posit that the reasoning process is implicitly modeled as a sequence of goal-setting and problem-solving steps; that is, the natural-language reasoning trace $\vz$ can be viewed as semantically equivalent to a sequence of question-answer pairs.
Formally, given a query $\vx$, we assume that for any reasoning-answer pair $(\vz,\vy)$, there exists a ground-truth decomposition $\mS_T \equiv (\vz,\vy)$ such that
\footnote{
Note that (i) the ground-truth decomposition may not be unique. E.g., $\{\vs_t\}_{t=1}^T$ and $\{\vs_t\}_{t=2}^T$ are both valid decompositions, with the latter representing a coarser process; and (ii) the true structure of the decomposition can be non-linear~\citep{teng2025atom}, though it can be mapped to a linear form in CoT reasoning.
}
\begin{align}
    \mS_T &= \{\vs_t \triangleq(\vq_t, \va_t)\}_{t\in[T]},
\end{align}
where each $\vs_t$ is a \emph{Socratic step}, $\va_T = \vy$ denotes the final answer, and the equivalence $\mS_T \equiv (\vz,\vy)$ implies that the oracle probability model $p$ satisfies
\begin{align}
    \label{eq:core}
    p(\vz,\vy \mid \vx) &= p(\{(\vq_t,\va_t)\}_{t \in [T]} \mid \vx).
\end{align}
Compared with the purely natural-language reasoning process $\vz$, the explicit sequence of Socratic steps offers clear advantages, most notably, finer-grained modeling and potential control of the reasoning process, enabling verification and intervention.
This explicit modeling lies at the heart of our proposed method, Socratic Self-Refine~(\ours), which we detail in \Secref{sec:method-ssr}.

\subsection{Socratic Self-Refine~(\ours): Decomposition, Verification, and Refinement}
\label{sec:method-ssr}

\textbf{From Entangled Reasoning to Explicit Socratic Process.}\quad
Under the assumption of \Eqref{eq:core}, our goal is to recover the full Socratic process $\mS_T$ from the natural-language reasoning trace $\vz$. Since no prior work explicitly models this process, and the oracle posterior $p(\mS_T \mid \vx,\vy,\vz)$ is unavailable, we adopt a zero-shot prompting approach with LLMs to decompose $\vz$ into the Socratic process $\mS_T$:
\begin{align}
    \label{eq:decompose}
    \mS_T \sim \pi_\vtheta(\cdot \mid \vx, \vy, \vz, \vx_{\text{dec}}) \approx p(\cdot \mid \vx, \vy, \vz),
\end{align}
where $\vx_{\text{dec}}$ denotes a decomposition query that prompts the LLM to extract a sequence of sub-questions and their corresponding sub-answers.
Leveraging prior work on LLM-based summarization and information extraction~\citep{van2024adapted}, this decomposition can be performed reliably with relatively little overhead.

\textbf{LLM Self-Verification on Socratic Steps.}\quad
We now leverage the reformulation of the original reasoning trace $\vz$ into the Socratic process $\mS_T$ to enable LLM self-verification.
The joint probability distribution of $\mS_T$ can be factorized into a product of conditional probabilities:
\begin{align}
    \label{eq:factorization}
    \pi_\vtheta(\mS_T \mid \vx)
    &=
    \pi_\vtheta(\{(\vq_t, \va_t)\}_{t\in[T]} \mid \vx)
    = \prod\nolimits_{t=1}^{T} 
    \underbrace{\pi_\vtheta(\vq_t \mid \{\vs_i\}_{i<t}, \vx)}_{\text{$t$-th step planning}} 
    \cdot 
    \underbrace{\pi_\vtheta(\va_t \mid \vq_t, \{\vs_i\}_{i<t}, \vx)}_{\text{$t$-th step execution}},
\end{align}
where $\{\vs_i\}_{i<1}\triangleq \emptyset$. 
This factorization captures our core perspective on LLM reasoning: at each step, the model first plans by formulating the next sub-question, and then executes by generating the corresponding sub-answer. Such a sequential formulation naturally lends itself to Monte Carlo search over possible reasoning trajectories, where the two types of actions are sub-question generation ($\vq$) and sub-answer generation ($\va$)~\citep{qi2024mutual,acuna2025socratic}. However, as the modern LLMs typically do not undergo the training of explicitly proposing and answering the next probable sub-questions, this approach might be less effective.

\ours evaluates the confidence of each sub-answer $\va_t$ given the current sub-question $\vq_t$, the original query $\vx$, and the context of the previous Socratic steps ${(\vq_i,\va_i)}_{i<t}$.
Specifically, we encode all relevant information into the context and ask the LLM to solve each sub-question independently $M$ times. The resulting answers form a reference set
\begin{align}
    \hat{\mA}_t = \{\hat{\va}_{ti}\}_{i\in[M]}, \quad \hat{\va}_{ti}\sim \pi_\vtheta(\cdot \mid \vq_t, \{\vs_i\}_{i<t}, \vx).
\end{align}
We then compare the original $t$-th step sub-answer $\va_t$ with $\hat{\mA}_t$ and estimate the confidence score as
\begin{align}
    \label{eq:conf}
    c_t = \frac{1}{M} \sum\nolimits_{i=1}^{M}\mathbbm{1}_{\va_t = \hat{\va}_{ti}}, \quad \forall t \in [T].
\end{align}
For mathematical problems, intermediate sub-answers can be restricted to mathematical expressions through simple prompting, allowing for deterministic equivalence checking. In practice, however, we find that this restriction does not consistently constrain LLM outputs
We therefore resort to LLM self-evaluation, producing confidence scores directly with a context-free confidence estimation prompt $\vx_{\text{conf}}$:
\begin{align}
    \mC_T &= \{c_t \sim \pi_\vtheta(\cdot \mid \va_t, \hat{\mA}_t, \vx_{\text{conf}})\}_{t\in [T]}.
\end{align}
These confidence scores are then used to guide refinement of the current-round reasoning and can also be aggregated to reflect overall response quality, which supports sample selection in our test-time scaling experiments~(\Secref{sec:experiments-tts}). 
It is worth noting that we enforce strict context management during confidence estimation: the prompt includes only the candidate sub-answer and the reference answer set, with no additional information. This design has two motivations. First, we assume that judging equivalence between expressions can be done in a context-free manner, i.e., with only the expressions. Second, isolating the context helps control the computation budget.

\textbf{LLM Self-Refinement with Socratic Steps.}\quad 
Once the confidence scores of all Socratic steps are estimated, we use them to guide reasoning refinement. In \ours, we first identify the step $t^\prime=\arg\min_{t}\{c_t\}_{t\in [T]}$ with the lowest confidence score $c_{t^\prime}$. We then apply majority voting over its reference answer set to obtain a refined sub-answer:
\begin{align}
    \va_{t^\prime}^* &= \arg\max_{\va} \pi_\vtheta(\va \mid \vq_{t^\prime}, \{\vs_i\}_{i<t^{\prime}},\vx)
    \approx \operatorname{maj\_vote}(\mA_{t^\prime}).
\end{align}
This refined sub-answer is injected into the iteration-$k$ refinement of $(\vz^{(k)}, \vy^{(k)})$, producing the next iteration:
\begin{align}
    (\vz^{(k+1)}, \vy^{(k+1)}) \sim \pi_{\vtheta}(\cdot \mid \vx, \vy^{(k)}, \vz^{(k)}, \underbrace{\vq^{(k)}_{t^\prime}, \va^{(k)}_{t^\prime}, \va^{*(k)}_{t^\prime}}_{\text{Socratic Feedback}}, \vx_{\text{ref}}), 
\end{align}
where the triplet $(\vq^{(k)}_{t^\prime}, \va^{(k)}_{t^\prime}, \va^{*(k)}_{t^\prime})$ is called Socratic Feedback, the template of which can be found in \appref{app:prompts}, and $\vx_{\text{ref}}$ is the refinement query that prompts the LLM to revise for a new reasoning trace $\vz^{(k+1)}$ that leads to $\va^{*(k)}_{t^\prime}$. 
Because most modern LLMs are trained with instruction tuning~\citep{wei2021finetuned} and preference tuning~\citep{ouyang2022training}, both relying on chain-of-thought-like structures, the direct injection of the Socratic process in unnatural formats (e.g., JSON) might disrupt reasoning.
Our design principle in \ours is therefore to minimize format  disruption and to inject only the necessary information into the refinement context. For a detailed analysis of this issue, please refer to \Secref{sec:experiments-ablation}.
\emph{We refer to the variant that directly combines the three steps described above as Linear \ours (\ours-Lin).}

\subsection{\ours Deployment: Better Efficiency and Beyond Step-Level Refinement}
\label{method:ssr-deployment}

\textbf{Improving the Efficiency of \ours with Gating Self-Refine.}\quad
Applying fine-grained, step-level \ours at every refinement step can be costly compared to other iterative refinement frameworks~\citep{madaan2023self,teng2025atom}.
To balance efficiency and accuracy, we adopt a gating mechanism that combines Self-Refine~\citep{madaan2023self} with \ours. 
In deployment, Self-Refine serves as the default refinement method, while \ours is invoked only when Self-Refine fails to identify mistakes in the reasoning trace or when the response is already correct. Because these two situations cannot be distinguished in advance, applying \ours in the latter case incurs only a minor additional cost, while in the former case it provides an extra layer of safety.
Compared to \ours only, this approach reduces overhead while preserving the advantages of \ours's ability of fine-grained step-level verification. 
\emph{We denote \ours with this adaptive gating mechanism as Adaptive \ours (\ours-Ada).}

\textbf{\ours Planning Refinement.}\quad
Our current \ours relies on two implicit assumptions about reasoning planning: (i) response quality evaluation is independent of high-level planning, and (ii) refinement focuses only on execution accuracy. These restrictions may limit the performance of \ours.
By assuming probabilistic independence between each sub-question $\vq_t$ and the preceding answers $\{\va_i\}_{i<t}$, the factorization
\footnote{Under this assumption, we posit that the LLM establishes an overall plan before generating the actual response~\citep{ye2024physics,lindsey2025biology}.} in \Eqref{eq:factorization} can be simplified as
\begin{equation}
\begin{aligned}
    \label{eq:plan}
    \pi_\vtheta(\mS_T \mid \vx)
    &=
    \underbrace{\pi_\vtheta(\{\vq_t\}_{t\in[T]} \mid \vx)\vphantom{\prod\nolimits_{t=1}^{T} 
    \pi_\vtheta(\va_t \mid \vq_t, \{\vs_i\}_{i<t}, \vx)}}_{\text{high-level planning}}
    \cdot
    \underbrace{\prod\nolimits_{t=1}^{T} 
    \pi_\vtheta(\va_t \mid \vq_t, \{\vs_i\}_{i<t}, \vx).}_{\text{sequential execution}}
\end{aligned}
\end{equation}
To ensure the reliability of high-level planning before applying step-level \ours, while keeping the overhead modest compared to other baselines, we perform only one round of plan refinement. Unlike our main \ours procedure, we do not repeatedly sample rollouts or evaluate their quality. Instead, we directly prompt the LLM to judge whether the high-level plan (a sequence of sub-questions or their natural-language description) is sufficiently sound for the subsequent execution.
\emph{We denote \ours-Ada plus this plan refinement as \ours-Plan.}
\textbf{For the detailed algorithmic description of our \ours, please refer to \Algref{alg:ssr} in the Appendix.}

\section{Experiments}
\label{sec:experiments}
We evaluate our \ours's effectiveness through comprehensive experiments, covering 
experimental setup~(\Secref{sec:experiments-setting}), 
main results on the mathematical and logical reasoning benchmarks~(\Secref{sec:experiments-main}),
ablation studies on the choice of incorporating Socratic content into refinement~(\Secref{sec:experiments-ablation}),
test-time scaling effect of our \ours~(\Secref{sec:experiments-tts}). 
{and granularity of the Socratic step decomposition~(\Secref{sec:experiments-granularity})}.
\textbf{For additional results, please refer to \appref{app:experiments}.}

\subsection{Settings}
\label{sec:experiments-setting}

\textbf{Models, Datasets, and Evaluation.}\quad 
We use the latest \gptnano~(general-purpose) and \gptfive~(reasoning) models from OpenAI as our LLM backbones, chosen for their balanced capabilities in instruction following and reasoning.
We additionally include the results {from another model family including \geminiflash and \geminilite~\citep{comanici2025gemini} in \appref{app:gemini}.} 

We benchmark the reasoning frameworks on two categories of datasets: \textbf{mathematical reasoning} and \textbf{logical reasoning}. For mathematical reasoning, we adopt the challenging Level-5 subset of the MATH dataset (\textbf{MATH-Level-5}) with numerical answers~\citep{hendrycks2021measuring}, American Invitational Mathematics Examination (\textbf{AIME}) from 2024 and 2025~\citep{aime}, and the math subset of Humanity's Last Exam~(HLE)~\citep{phan2025humanity}. We adopt the library of Math-Verify~\citep{math-verify} for matching the candidate and ground-truth answer (except for the non-numerical subset of HLE).
For logical reasoning, we use the synthetic reasoning-gym environment~\citep{stojanovski2025reasoning} to generate sub-tasks including the Zebra Puzzle and Mini-Sudoku{, where we use exact string matching and rule-based verifier as the evaluation, respectively.}

\textbf{Baselines.}\quad 
We compare our \ours against several iterative refinement-based test-time LLM reasoning frameworks. 
\textbf{Self-Refine}~\citep{madaan2023self} iteratively generates feedback for a given response and updates the response based on this self-feedback. 
{\textbf{Debate}~\citep{du2023improving} employs a multi-agent framework in which each agent iteratively refines or defends its response by engaging with the responses of peer agents.}
Monte Carlo Tree Self-Refine (\textbf{MCTSr})~\citep{zhang2024accessing} treats the full generation as a node and the self-refine step as an edge, applying Monte Carlo Tree Search~(MCTS) to search for the best response. 
Atom of Thoughts~(\textbf{AoT})~\citep{teng2025atom} incrementally constructs a Directed Acyclic Graph (DAG) of reasoning, contracts intermediate results into improved sub-questions, and solves them step by step. 
We do not include parallel sampling-based baselines such as Forest of Thoughts~(\textbf{FoT})~\citep{bi2024forest}, since these approaches are complementary to iterative refinement methods. Their benefits are instead reflected through the \textbf{Maj@5} metric in \Tabref{tab:main}.

\textbf{Implementation of \ours.}\quad 
We implement and evaluate three variants of \ours in \Secref{method:ssr-deployment}. Linear \ours (\textbf{\ours-Lin}) applies Socratic self-refine at every iteration, making it the most costly but also the most thorough approach to step-level fine-grained refinement.
Adaptive \ours (\textbf{\ours-Ada}) first applies the basic Self-Refine; if the feedback reveals clear and critical errors, the feedback is directly adopted, while if no errors are detected, the method falls back to Socratic self-refine. \ours with plan refinement (\textbf{\ours-Plan})  adds an initial round of plan refinement before the step-level Socratic self-refine, thereby equipping \ours with high-level refinement capabilities. 
\textbf{For more details, please refer to \appref{app:baselines}.}

\subsection{\ours's Step-Level Verification Leads to Consistent Performance Gains}
\label{sec:experiments-main}
\textbf{\Tabref{tab:main} and \Tabref{tab:main-pass-at-k} show results on comprehensive metrics for various methods.}

Overall, the proposed \ours variants bring substantial improvements when powered by the strong \gptfive. Across all tasks, \ours consistently surpasses competitive baselines, yielding clear gains in both LR-Acc and LR-Maj@5. Notably, \ours-Plan achieves the best or second-best results in nearly every setting, with particularly large margins on challenging mathematical reasoning benchmarks like AIME. This highlights that structured preliminary planning amplifies the benefits of iterative refinement, even when starting from already strong \gptfive reasoning capabilities. Our framework also demonstrates effectiveness on the weaker \gptnano backbone. Despite its limited reasoning capacity, all three \ours variants in general improve performance over baselines, underscoring that our refinement strategies generalize across model scales. This implies a viable path of adopting our \ours to boost smaller, resource-efficient models.

\begin{table*}[t]
\caption{
    \textbf{Last-Round Performance of Iterative Refinement-Based Reasoning Methods.} 
    \textbf{LR-Acc:} Last-round refinement's accuracy, yielded by 10 repeated experiments;
    \textbf{LR-Maj@5:} Last-round refinement's accuracy of majority voting with 5 samples in parallel, yielded by 50 repeated experiments.
    \textbf{Boldface} and \underline{underlining} denote the best and the second-best performance, respectively. 
}
\vspace{-1.2em}
\begin{center}
\resizebox{1\linewidth}{!}{%
\setlength{\tabcolsep}{2pt}
\begin{tabular}{l cc cc cc cc cc }
	\toprule[0.12em] 
    \multirow{2}{*}[-0.25em]{\textbf{Method}} 
    & \multicolumn{2}{c}{\textbf{MATH-Level-5}}
    & \multicolumn{2}{c}{\textbf{AIME24}} 
    & \multicolumn{2}{c}{\textbf{AIME25}} 
    & \multicolumn{2}{c}{\textbf{Zebra-Puzzle}}
    & \multicolumn{2}{c}{\textbf{Mini-Sudoku}}
    \\
    \cmidrule(lr){2-3} \cmidrule(lr){4-5} \cmidrule(lr){6-7} \cmidrule(lr){8-9} \cmidrule(lr){10-11} 
    % \cmidrule(lr){12-13} 
     
    & \emph{LR-Acc} & \emph{LR-Maj@5}
    & \emph{LR-Acc} & \emph{LR-Maj@5}
    & \emph{LR-Acc} & \emph{LR-Maj@5}
    & \emph{LR-Acc} & \emph{LR-Maj@5}
    & \emph{LR-Acc} & \emph{LR-Maj@5}
    \\
     
    \midrule

    %%%%%%%%%%%%%%%%%%%%%%%%%%%%%%%%%%%%
    %%%%%%%%%%%% GPT-4.1-nano
    %%%%%%%%%%%%%%%%%%%%%%%%%%%%%%%%%%%%
    
    \multicolumn{11}{c}{\gptnano} 
    \\
    \midrule

    CoT
    & 74.88\scriptsize{$\pm$1.35}
    & 82.32\scriptsize{$\pm$1.11}
    & 27.00\scriptsize{$\pm$4.58}
    & 32.80\scriptsize{$\pm$2.15}
    & 23.00\scriptsize{$\pm$3.48}
    & 26.93\scriptsize{$\pm$2.97}
    & \underline{55.20\scriptsize{$\pm$3.28}}
    & \underline{56.56\scriptsize{$\pm$2.44}}
    & 47.40\scriptsize{$\pm$3.35}
    & 66.04\scriptsize{$\pm$2.69}

    \\

    Self-Refine
    & 68.69\scriptsize{$\pm$1.15}
    & 79.81\scriptsize{$\pm$0.75}
    & 28.00\scriptsize{$\pm$4.99}
    & 34.33\scriptsize{$\pm$3.00}
    & 22.67\scriptsize{$\pm$2.91}
    & 28.33\scriptsize{$\pm$3.42}
    & 53.50\scriptsize{$\pm$1.96}
    & {56.08\scriptsize{$\pm$1.93}}
    & \underline{53.60\scriptsize{$\pm$4.59}}
    & 73.04\scriptsize{$\pm$3.21}
    \\

    {Debate}
    & \textbf{79.28\scriptsize{$\pm$0.86}}
    & \textbf{84.08\scriptsize{$\pm$0.76}}
    & 27.00\scriptsize{$\pm$4.82}
    & 32.40\scriptsize{$\pm$3.13}
    & \textbf{26.67\scriptsize{$\pm$2.58}}
    & \underline{27.60\scriptsize{$\pm$2.75}}
    & 54.70\scriptsize{$\pm$3.29}
    & 57.16\scriptsize{$\pm$2.66}
    & \textbf{60.80\scriptsize{$\pm$4.81}}
    & \textbf{78.38\scriptsize{$\pm$2.75}}
    \\

    MCTSr %\scriptsize{(d=4)}
    & 74.02\scriptsize{$\pm$1.12}
    & 83.01\scriptsize{$\pm$0.81}
    & 23.67\scriptsize{$\pm$4.33}
    & 30.47\scriptsize{$\pm$3.13}
    & 20.00\scriptsize{$\pm$4.94}
    & 25.73\scriptsize{$\pm$4.22}
    & {54.90\scriptsize{$\pm$2.47}}
    & 54.88\scriptsize{$\pm$2.45}
    & {53.33\scriptsize{$\pm$1.63}}
    & \underline{73.84\scriptsize{$\pm$2.43}}
    \\

    AoT
    & 75.15\scriptsize{$\pm$1.00}
	& 82.83\scriptsize{$\pm$0.83}
	& 21.11\scriptsize{$\pm$4.97}
	& 25.67\scriptsize{$\pm$3.61}
	& 21.33\scriptsize{$\pm$3.06}
	& 25.53\scriptsize{$\pm$3.75}
	& 29.33\scriptsize{$\pm$3.16}
	& 43.60\scriptsize{$\pm$2.65}
	& 42.80\scriptsize{$\pm$2.96}
	& 65.08\scriptsize{$\pm$2.26}
    \\

    \rowcolor{lightergray}
    \ours-Lin~(Ours)
    & \underline{77.06\scriptsize{$\pm$0.93}}
    & {83.64\scriptsize{$\pm$0.69}}
    & \textbf{32.67\scriptsize{$\pm$3.59}}
    & \textbf{39.93\scriptsize{$\pm$3.23}}
    & {24.00\scriptsize{$\pm$4.67}}
    & 27.33\scriptsize{$\pm$4.06}
    & 54.60\scriptsize{$\pm$2.20}
    & 54.10\scriptsize{$\pm$2.09}
    & 53.10\scriptsize{$\pm$2.47}
    & 72.76\scriptsize{$\pm$2.55}
    \\

    \rowcolor{lightergray}
    \ours-Ada~(Ours)
    & 75.70\scriptsize{$\pm$1.31}
    & 82.71\scriptsize{$\pm$0.90}
    & \underline{29.67\scriptsize{$\pm$6.74}}
    & \underline{37.47\scriptsize{$\pm$4.25}}
    & \underline{24.67\scriptsize{$\pm$3.06}}
    & \textbf{28.80\scriptsize{$\pm$3.38}}
    & 54.30\scriptsize{$\pm$1.90}
    & 55.14\scriptsize{$\pm$1.71}
    & 51.50\scriptsize{$\pm$4.41}
    & {73.22\scriptsize{$\pm$3.37}}
    \\

    \rowcolor{lightergray}
    \ours-Plan~(Ours)
    & {76.01\scriptsize{$\pm$0.57}}
    & \underline{83.75\scriptsize{$\pm$0.74}}
    & 27.33\scriptsize{$\pm$5.73}
    & 35.80\scriptsize{$\pm$3.39}
    & 22.33\scriptsize{$\pm$3.67}
    & 27.53\scriptsize{$\pm$4.46}
    & \textbf{56.90\scriptsize{$\pm$3.11}}
    & \textbf{57.30\scriptsize{$\pm$2.39}}
    & 47.70\scriptsize{$\pm$4.22}
    & 66.46\scriptsize{$\pm$4.61}
    \\

    \midrule

    %%%%%%%%%%%%%%%%%%%%%%%%%%%%%%%%%%%%
    %%%%%%%%%%%% GPT-5-mini
    %%%%%%%%%%%%%%%%%%%%%%%%%%%%%%%%%%%%
    
    \multicolumn{11}{c}{\gptfive} \\

    \midrule

    CoT
    & 82.95\scriptsize{$\pm$1.02}
    & 90.05\scriptsize{$\pm$0.54}
    & 50.67\scriptsize{$\pm$4.67}
    & 60.87\scriptsize{$\pm$3.93}
    & 37.00\scriptsize{$\pm$6.57}
    & 49.80\scriptsize{$\pm$4.19}
    & 82.80\scriptsize{$\pm$2.71}
    & 91.00\scriptsize{$\pm$1.30}
    & 42.40\scriptsize{$\pm$2.42}
    & 61.96\scriptsize{$\pm$3.19}
    \\

    Self-Refine
    & 87.02\scriptsize{$\pm$1.40}
    & 94.11\scriptsize{$\pm$0.47}
    & 63.33\scriptsize{$\pm$4.94}
    & 74.40\scriptsize{$\pm$3.74}
    & 53.67\scriptsize{$\pm$6.23}
    & 68.33\scriptsize{$\pm$3.48}
    & 82.00\scriptsize{$\pm$2.61}
    & 92.64\scriptsize{$\pm$1.61}
    & 63.60\scriptsize{$\pm$3.35}
    & 93.82\scriptsize{$\pm$1.35}
    \\

    {Debate} 
    & 90.62\scriptsize{$\pm$0.94}
    & 93.47\scriptsize{$\pm$0.46}
    & 63.67\scriptsize{$\pm$3.79}
    & 74.13\scriptsize{$\pm$3.44}
    & 53.33\scriptsize{$\pm$3.33}
    & 61.87\scriptsize{$\pm$3.21}
    & \textbf{91.20\scriptsize{$\pm$1.72}}
    & \textbf{93.74\scriptsize{$\pm$1.07}}
    & 90.40\scriptsize{$\pm$3.95}
    & 98.54\scriptsize{$\pm$1.31}
    \\

    MCTSr %\scriptsize{(d=4)}
    & {87.42\scriptsize{$\pm$0.89}}
    & 92.91\scriptsize{$\pm$0.71}
    & 57.00\scriptsize{$\pm$5.67}
    & 68.87\scriptsize{$\pm$4.35}
    & 46.97\scriptsize{$\pm$6.11}
    & 55.40\scriptsize{$\pm$4.76}
    & 83.00\scriptsize{$\pm$1.90}
    & 89.82\scriptsize{$\pm$1.49}
    & 61.40\scriptsize{$\pm$6.17}
    & 89.68\scriptsize{$\pm$2.56}
    \\

    AoT
    & 80.56\scriptsize{$\pm$0.63}
	& 88.84\scriptsize{$\pm$0.60}
	& 46.67\scriptsize{$\pm$5.16}
	& 57.00\scriptsize{$\pm$3.21}
	& 33.00\scriptsize{$\pm$6.05}
	& 43.60\scriptsize{$\pm$3.82}
	& 65.30\scriptsize{$\pm$3.07}
	& 74.78\scriptsize{$\pm$2.07}
	& 61.70\scriptsize{$\pm$3.72}
	& 82.72\scriptsize{$\pm$2.75}
    \\

    \rowcolor{lightergray}
    \ours-Lin~(Ours)
    & {88.36\scriptsize{$\pm$1.06}}
    & {93.01\scriptsize{$\pm$0.63}}
    & {64.00\scriptsize{$\pm$5.12}}
    & {74.60\scriptsize{$\pm$4.10}}
    & {55.67\scriptsize{$\pm$4.48}}
    & {65.47\scriptsize{$\pm$3.76}}
    & {87.70\scriptsize{$\pm$2.97}}
    & \underline{93.70\scriptsize{$\pm$1.76}}
    & {93.60\scriptsize{$\pm$1.69}}
    & {99.70\scriptsize{$\pm$0.54}}
    \\

    \rowcolor{lightergray}
    \ours-Ada~(Ours)
    & \underline{91.57\scriptsize{$\pm$0.51}}
    & \underline{95.62\scriptsize{$\pm$0.35}}
    & \underline{68.67\scriptsize{$\pm$4.52}}
    & \underline{75.93\scriptsize{$\pm$3.08}}
    & \underline{60.33\scriptsize{$\pm$4.58}}
    & \underline{70.13\scriptsize{$\pm$3.46}}
    & {87.30\scriptsize{$\pm$2.53}}
    & 93.00\scriptsize{$\pm$1.69}
    & \textbf{96.10\scriptsize{$\pm$2.07}}
    & \underline{99.98\scriptsize{$\pm$0.14}}
    \\

    \rowcolor{lightergray}
    \ours-Plan~(Ours)
    & \textbf{92.16\scriptsize{$\pm$0.67}}
    & \textbf{95.93\scriptsize{$\pm$0.30}}
    & \textbf{69.67\scriptsize{$\pm$4.82}}
    & \textbf{79.00\scriptsize{$\pm$3.48}}
    & \textbf{62.00\scriptsize{$\pm$6.18}}
    & \textbf{71.53\scriptsize{$\pm$5.26}}
    & \underline{88.00\scriptsize{$\pm$1.55}}
    & {93.20\scriptsize{$\pm$1.08}}
    & \underline{94.80\scriptsize{$\pm$2.48}}
    & \textbf{100.00\scriptsize{$\pm$0.00}}
    \\
    
    \bottomrule[0.12em]
    \end{tabular}
}
\end{center}
\label{tab:main}
\vspace{-0em}
\end{table*}

Second, the results in Table 2 show that \ours maintains superiority under upper-bound evaluation metrics. Both BoK-Acc and Pass@K demonstrate that \ours variants yield higher-quality and diverse refinement trajectories compared to baselines. Again, \ours-Plan often achieves the best results, while \ours-Ada provides a favorable trade-off between efficiency and accuracy, confirming the value of adaptively combining Self-Refine with Socratic refinement.

Finally, the comparison across reasoning categories highlights complementary strengths. In mathematical reasoning, \ours gains from explicit verification and refinement of sub-answers, which reduces cascading errors in long derivations. In logical reasoning tasks such as Zebra-Puzzle and Mini-Sudoku, where execution accuracy dominates, step-level Socratic verification also proves highly effective, often yielding substantial improvements over baselines.

Overall, the experiments confirm that the explicit modeling and verification of Socratic steps in \ours provides more reliable and controllable refinement than existing iterative approaches, with \ours-Plan standing out as the most robust variant.

\begin{table*}[t]
\caption{
    \textbf{Upper-Bound Performance of Iterative Refinement-Based Reasoning Methods.} 
    \textbf{BoK-Acc:} Best-of-K refinements' accuracy, yielded by prompting LLM-as-a-Judge~\citep{gu2024survey} for selecting the best answer out of K iterations of refinement;
    \textbf{Pass@K:} Pass-at-K refinements' accuracy (at lease one of K iterations gets the answer correct).
    Both experiments are repeated for 10 times.
    \textbf{Boldface} and \underline{underlining} denote the best and the second-best performance, respectively. 
}
\vspace{-1.2em}
\begin{center}
\resizebox{1\linewidth}{!}{%
\setlength{\tabcolsep}{4pt}
\begin{tabular}{l cc cc cc cc cc }
	\toprule[0.12em] 
    \multirow{2}{*}[-0.25em]{\textbf{Method}} 
    & \multicolumn{2}{c}{\textbf{MATH-Level-5}}
    & \multicolumn{2}{c}{\textbf{AIME24}} 
    & \multicolumn{2}{c}{\textbf{AIME25}} 
    & \multicolumn{2}{c}{\textbf{Zebra-Puzzle}}
    & \multicolumn{2}{c}{\textbf{Mini-Sudoku}}
    \\
    \cmidrule(lr){2-3} \cmidrule(lr){4-5} \cmidrule(lr){6-7} \cmidrule(lr){8-9} \cmidrule(lr){10-11} 
     
    & \emph{BoK-Acc} & \emph{Pass@K}
    & \emph{BoK-Acc} & \emph{Pass@K}
    & \emph{BoK-Acc} & \emph{Pass@K}
    & \emph{BoK-Acc} & \emph{Pass@K}
    & \emph{BoK-Acc} & \emph{Pass@K}
    \\
     
    \midrule

    %%%%%%%%%%%%%%%%%%%%%%%%%%%%%%%%%%%%
    %%%%%%%%%%%% GPT-4.1-nano
    %%%%%%%%%%%%%%%%%%%%%%%%%%%%%%%%%%%%
    
    \multicolumn{11}{c}{\gptnano} 
    \\
    \midrule

    CoT
    & \multicolumn{1}{c}{74.88\scriptsize{$\pm$1.35}}
    & -
    & \multicolumn{1}{c}{27.00\scriptsize{$\pm$4.58}}
    & -
    & \multicolumn{1}{c}{23.00\scriptsize{$\pm$3.48}}
    & -
    & \multicolumn{1}{c}{\underline{55.20\scriptsize{$\pm$3.28}}}
    & -
    & \multicolumn{1}{c}{47.40\scriptsize{$\pm$3.35}}
    & -
    \\

    Self-Refine
    & 76.48\scriptsize{$\pm$0.95}
    & 81.60\scriptsize{$\pm$0.82}
    & 30.67\scriptsize{$\pm$5.54}
    & 31.67\scriptsize{$\pm$5.00}
    & 23.67\scriptsize{$\pm$4.07}
    & 26.00\scriptsize{$\pm$4.90}
    & 55.60\scriptsize{$\pm$3.77}
    & \underline{59.60\scriptsize{$\pm$2.37}}
    & 56.90\scriptsize{$\pm$5.84}
    & 65.70\scriptsize{$\pm$3.55}
    \\

    {Debate}
    & \textbf{79.62\scriptsize{$\pm$0.79}}
    & 84.51\scriptsize{$\pm$1.01}
    & 29.00\scriptsize{$\pm$3.00}
    & 35.33\scriptsize{$\pm$3.40}
    & \underline{26.00\scriptsize{$\pm$3.89}}
    & 31.00\scriptsize{$\pm$3.67}
    & \textbf{56.80\scriptsize{$\pm$2.79}}
    & \textbf{68.50\scriptsize{$\pm$4.06}}
    & \textbf{63.50\scriptsize{$\pm$3.96}}
    & 70.70\scriptsize{$\pm$3.44}
    \\

    AoT
    & \underline{79.37\scriptsize{$\pm$1.54}}
    & \textbf{87.28\scriptsize{$\pm$0.64}}
    & 23.33\scriptsize{$\pm$5.21}
    & 33.70\scriptsize{$\pm$3.99}
    & 24.33\scriptsize{$\pm$4.48}
    & 29.33\scriptsize{$\pm$5.33}
    & 37.33\scriptsize{$\pm$3.20}
    & 63.22\scriptsize{$\pm$3.64}
    & 50.20\scriptsize{$\pm$5.08}
    & \textbf{76.00\scriptsize{$\pm$3.26}}
    \\

    \rowcolor{lightergray}
    \ours-Lin~(Ours)
    & 78.03\scriptsize{$\pm$1.00}
    & 82.97\scriptsize{$\pm$0.98}
    & \textbf{33.33\scriptsize{$\pm$4.22}}
    & \textbf{38.33\scriptsize{$\pm$5.63}}
    & \textbf{26.67\scriptsize{$\pm$3.94}}
    & \underline{32.00\scriptsize{$\pm$4.00}}
    & \underline{55.90\scriptsize{$\pm$2.74}}
    & \underline{65.40\scriptsize{$\pm$1.96}}
    & \underline{58.20\scriptsize{$\pm$3.71}}
    & \underline{75.40\scriptsize{$\pm$3.38}}
    \\

    \rowcolor{lightergray}
    \ours-Ada~(Ours)
    & 78.05\scriptsize{$\pm$1.37}
    & 85.14\scriptsize{$\pm$0.56}
    & \underline{31.67\scriptsize{$\pm$5.82}}
    & \underline{36.33\scriptsize{$\pm$5.67}}
    & {25.67\scriptsize{$\pm$4.48}}
    & 32.00\scriptsize{$\pm$3.40}
    & 55.30\scriptsize{$\pm$1.19}
    & 62.80\scriptsize{$\pm$2.04}
    & 56.70\scriptsize{$\pm$3.44}
    & 74.20\scriptsize{$\pm$4.94}
    \\

    \rowcolor{lightergray}
    \ours-Plan~(Ours)
    & {78.40\scriptsize{$\pm$1.10}}
    & \underline{85.27\scriptsize{$\pm$0.47}}
    & 31.33\scriptsize{$\pm$5.42}
    & 35.67\scriptsize{$\pm$4.23}
    & 24.33\scriptsize{$\pm$3.67}
    & \textbf{34.33\scriptsize{$\pm$5.17}}
    & \underline{56.60\scriptsize{$\pm$3.58}}
    & \underline{64.60\scriptsize{$\pm$3.01}}
    & 56.40\scriptsize{$\pm$4.05}
    & 73.70\scriptsize{$\pm$2.37}
    \\

    \midrule

    %%%%%%%%%%%%%%%%%%%%%%%%%%%%%%%%%%%%
    %%%%%%%%%%%% GPT-5-mini
    %%%%%%%%%%%%%%%%%%%%%%%%%%%%%%%%%%%%
    
    \multicolumn{11}{c}{\gptfive} \\

    \midrule

    CoT
    & \multicolumn{1}{c}{82.95\scriptsize{$\pm$1.02}}
    & -
    & \multicolumn{1}{c}{50.67\scriptsize{$\pm$4.67}}
    & -
    & \multicolumn{1}{c}{37.00\scriptsize{$\pm$6.57}}
    & -
    & \multicolumn{1}{c}{82.80\scriptsize{$\pm$2.71}}
    & -
    & \multicolumn{1}{c}{42.40\scriptsize{$\pm$2.42}}
    & -
    \\

    Self-Refine
    & 89.40\scriptsize{$\pm$1.00}
    & 91.59\scriptsize{$\pm$0.83}
    & 61.33\scriptsize{$\pm$4.00}
    & 68.00\scriptsize{$\pm$3.71}
    & 51.67\scriptsize{$\pm$6.87}
    & 56.67\scriptsize{$\pm$6.67}
    & 90.90\scriptsize{$\pm$2.21}
    & 91.30\scriptsize{$\pm$1.79}
    & 85.70\scriptsize{$\pm$3.23}
    & 83.30\scriptsize{$\pm$2.19}
    \\

    {Debate}
    & 90.43\scriptsize{$\pm$0.88}
    & 91.70\scriptsize{$\pm$0.79}
    & 64.00\scriptsize{$\pm$4.16}
    & 64.67\scriptsize{$\pm$4.27}
    & 53.00\scriptsize{$\pm$2.77}
    & 55.00\scriptsize{$\pm$2.69}
    & 91.70\scriptsize{$\pm$1.62}
    & 93.70\scriptsize{$\pm$1.35}
    & 90.20\scriptsize{$\pm$3.54}
    & 91.80\scriptsize{$\pm$3.57}
    \\

    AoT
    & 85.87\scriptsize{$\pm$0.49}
    & 91.38\scriptsize{$\pm$0.80}
    & 56.67\scriptsize{$\pm$6.15}
    & 61.67\scriptsize{$\pm$5.82}
    & 39.33\scriptsize{$\pm$3.27}
    & 49.00\scriptsize{$\pm$5.39}
    & 88.80\scriptsize{$\pm$1.94}
    & \textbf{93.50\scriptsize{$\pm$1.43}}
    & 93.70\scriptsize{$\pm$1.73}
    & 90.70\scriptsize{$\pm$2.15}
    \\

    \rowcolor{lightergray}
    \ours-Lin~(Ours)
    & 88.16\scriptsize{$\pm$1.31}
    & 89.54\scriptsize{$\pm$1.25}
    & 65.33\scriptsize{$\pm$5.42}
    & 67.00\scriptsize{$\pm$3.79}
    & 55.33\scriptsize{$\pm$7.02}
    & 59.00\scriptsize{$\pm$5.17}
    & \underline{92.20\scriptsize{$\pm$2.23}}
    & 93.20\scriptsize{$\pm$2.60}
    & 95.30\scriptsize{$\pm$1.19}
    & 95.50\scriptsize{$\pm$1.57}
    \\

    \rowcolor{lightergray}
    \ours-Ada~(Ours)
    & \underline{93.14\scriptsize{$\pm$0.52}}
    & \underline{94.63\scriptsize{$\pm$0.36}}
    & \textbf{71.67\scriptsize{$\pm$4.28}}
    & \textbf{74.00\scriptsize{$\pm$4.90}}
    & \underline{61.00\scriptsize{$\pm$4.73}}
    & \underline{66.00\scriptsize{$\pm$3.89}}
    & 91.80\scriptsize{$\pm$1.89}
    & 93.00\scriptsize{$\pm$1.84}
    & \underline{98.20\scriptsize{$\pm$1.25}}
    & \underline{98.10\scriptsize{$\pm$1.45}}
    \\

    \rowcolor{lightergray}
    \ours-Plan~(Ours)
    & \textbf{93.48\scriptsize{$\pm$0.52}}
    & \textbf{95.05\scriptsize{$\pm$0.34}}
    & \underline{71.00\scriptsize{$\pm$4.48}}
    & \underline{73.67\scriptsize{$\pm$4.07}}
    & \textbf{65.67\scriptsize{$\pm$6.16}}
    & \textbf{69.67\scriptsize{$\pm$5.26}}
    & \textbf{92.30\scriptsize{$\pm$1.62}}
    & \underline{93.30\scriptsize{$\pm$1.79}}
    & \textbf{98.70\scriptsize{$\pm$1.00}}
    & \textbf{98.30\scriptsize{$\pm$1.19}}
    \\
    
    \bottomrule[0.12em]
    \end{tabular}
}
\end{center}
\label{tab:main-pass-at-k}
\vspace{-0em}
\end{table*}

\subsection{{When Self-Refine Breaks, \ours Thrives: Extending \ours to Challenging Tasks}}
\label{sec:experiments-hle}
\begin{table}[t]
\caption{
Accuracies (\%) of iterative refinement-based reasoning methods on the 915-question text-only math subset of Humanity’s Last Exam~(\textbf{HLE})~\citep{phan2025humanity}, with \gptfive and \texttt{GPT-5} (medium reasoning, medium verbosity).
}
\vspace{-0.6em}
\centering
\resizebox{0.7\linewidth}{!}{
\setlength{\tabcolsep}{17pt}
\begin{tabular}{lccc}
    \toprule[0.12em]
    \textbf{Model}
    & \textbf{CoT}
    & \textbf{Self-Refine}
    & \textbf{\ours-Plan~(Ours)}
    \\
    \midrule 

    \gptfive 
    & 16.18
    & 18.58~{\textcolor{ForestGreen}{(+2.40)}}
    & \textbf{21.53~{\textcolor{ForestGreen}{(+5.35)}}}
    \\
     
    \texttt{GPT-5}
    & 27.98
    & 26.57~{\textcolor{Red}{(-1.41)}}
    & \textbf{29.61~{\textcolor{ForestGreen}{(+1.63)}}}
    \\
     
    \bottomrule[0.12em]
    \end{tabular}
 }
\label{tab:hle}
\vspace{0.0em}
\end{table}

In this section, we evaluate the effectiveness of \ours using more recent and stronger models, which require more challenging tasks to avoid performance saturation. Specifically, we employ the full \texttt{GPT-5} model in medium reasoning and medium verbosity modes, \emph{without tool calling or web searching}, and conduct experiments on Humanity’s Last Exam~(\textbf{HLE})~\citep{phan2025humanity}.
Due to budget constraints, we restrict our evaluation to the 915-question text-only math subset of HLE, where all questions are purely textual. We further divide this subset into two partitions based on whether the ground-truth answers are numerical. For the 478-example numerical partition, we follow the Math-Verify~\citep{math-verify} evaluation protocol described above, while for the 437-example non-numerical partition, we adopt the official LLM-as-a-Judge evaluation protocol with \texttt{GPT-5}.
The remaining settings are kept identical to those described earlier. \textbf{See \appref{app:hle} for details.}

\textbf{The results are reported in \Tabref{tab:hle}.}
our \ours framework consistently outperforms both Chain-of-Thought (CoT) and Self-Refine baselines across model scales. With \texttt{GPT-5-mini}, \ours achieves 21.53\% accuracy, surpassing CoT by 5.35 points and Self-Refine by 2.95 points, indicating that our two-level refinement reasoning framework is particularly beneficial for smaller models with limited reasoning capability. When scaled to the full \texttt{GPT-5}, \ours still yields a gain of 3.04 points over Self-Refine and 1.63 over CoT, suggesting that our approach complements intrinsic reasoning abilities rather than relying on model size alone.
Notably, it remains effective even for \texttt{GPT-5} where vanilla Self-Refine fails to generalize.
These results confirm that \ours effectively enhances iterative reasoning robustness for stronger frontier models like \texttt{GPT-5} even in challenging tasks such as HLE.

\subsection{Analysis: \ours Context Management}
\label{sec:experiments-ablation}
\begin{wrapfigure}{R}{0.6\textwidth}
\noindent\begin{minipage}{\linewidth}
\begin{table}[H]
\vspace{-\baselineskip}
\caption{
    \textbf{Ablation Study on \ours Context Management,} evaluated on \gptfive.
}
\vspace{-0.6em}
\centering
\resizebox{1\linewidth}{!}{%
\setlength{\tabcolsep}{7pt}
\begin{tabular}{lcc cc}
    \toprule[0.12em]
    \multirow{2}{*}[-0.25em]{\textbf{Method}} 
    & \multirow{2}{*}[-0.25em]{\textbf{Refinement}}
    & \multirow{2}{*}[-0.25em]{\textbf{Context}} 
    & \multicolumn{2}{c}{\textbf{Dataset}}
    \\
    \cmidrule(lr){4-5}

    & & 
    & AIME24
    & AIME25
    \\
    \midrule

    CoT & - & -
    & 50.67\scriptsize{$\pm$4.67}
    & 37.00\scriptsize{$\pm$6.57}
    \\

    \midrule 
     
    Self-Refine & Reflection & Natural
    & 63.33\scriptsize{$\pm$4.94}
    & 53.67\scriptsize{$\pm$6.23}
    \\

    \midrule

    \multirow{4}{*}[-0.17em]{\makecell{\ours-Plan\\(Ours)}} 
    &\cellcolor{lightergray} Reflection 
    &\cellcolor{lightergray} Natural
    &\cellcolor{lightergray}\textbf{69.67\scriptsize{$\pm$4.82}}
    &\cellcolor{lightergray}\textbf{62.00\scriptsize{$\pm$6.18}}
    \\

    &\cellcolor{lightergray} Reflection 
    &\cellcolor{lightergray} Socratic
    &\cellcolor{lightergray}\underline{67.67\scriptsize{$\pm$4.48}}
    &\cellcolor{lightergray}\underline{60.33\scriptsize{$\pm$4.82}}
    \\

    &\cellcolor{lightergray} Intervention 
    &\cellcolor{lightergray} Natural
    &\cellcolor{lightergray}54.67\scriptsize{$\pm$4.76}
    &\cellcolor{lightergray}42.67\scriptsize{$\pm$7.12}
    \\

    &\cellcolor{lightergray} Intervention 
    &\cellcolor{lightergray} Socratic
    &\cellcolor{lightergray}57.00\scriptsize{$\pm$8.09}
    &\cellcolor{lightergray}52.00\scriptsize{$\pm$5.62}
    \\
     
    \bottomrule[0.12em]
    \end{tabular}
    
 }
\label{tab:ablation}
\vspace{-1em}
\end{table}
\end{minipage}
\end{wrapfigure}
As discussed in \Secref{sec:method-ssr}, representing a natural language reasoning trace $\vz$ as a Socratic process $\mS_T$ requires careful consideration, since it introduces a distributional shift between the model’s training data and our artificially structured context. In this subsection, we explore alternative ways of integrating the Socratic process $\mS_T$ into reasoning refinement. Specifically, we focus on two key aspects:

\begin{itemize}[nosep,leftmargin=24pt]
    \item \textbf{Context Format}~\emph{(Natural / Socratic)}: Iterative refinement can be performed using only the Socratic steps $\mS_T$ \emph{(Socratic)}, discarding the original natural language reasoning trace $\vz$; or conversely, using only $\vz$ without the Socratic decomposition \emph{(Natural)}.  
    \item \textbf{Context Completeness}~\emph{(Reflection / Intervention)}: Since LLM chain-of-thought reasoning assumes linear dependencies, once the first problematic step $\vs_{t^{\prime}}$ is identified, later steps can be discarded. Refinement may then intervene directly at the error location \emph{(Intervention)}, avoiding unnecessary tokens, unlike \ours which refines after the full reasoning is completed \emph{(Reflection)}.  
\end{itemize}
\textbf{The results are reported in \Tabref{tab:ablation}.}
From the table, we observe that our implementation adopted in the main experiments (\emph{reflection} + \emph{natural context}) yields the strongest results (69.67 on AIME24 and 62.00 on AIME25), outperforming both Self-Refine and other variants of \ours. This suggests that \emph{preserving the original reasoning trace while applying reflection-based precise step-level refinement provides the model with richer contextual cues for error correction.}

Under reflection, replacing the natural context with the Socratic context yields slightly weaker but still competitive results, suggesting that while Socratic decomposition supports step-level analysis, it may miss some nuances of natural language reasoning. In contrast, intervention-based refinement consistently underperforms, as prematurely truncating the reasoning trace discards useful contextual information and leads to weaker refinements.

\subsection{Analysis: Test-Time Scaling of \ours}
\label{sec:experiments-tts}
In this subsection, we investigate whether the performance gains of \ours can be sustained under increased test-time compute. Test-time scaling for iterative refinement generally follows two orthogonal approaches: \emph{(i)~sequential scaling}, which increases the number of refinement iterations, and \emph{(ii)~parallel scaling}, which runs multiple refinements in parallel and aggregates the outputs.

\begin{wrapfigure}{r}{0.65\textwidth}
    \vspace{-0.8em}
    \centering
    \includegraphics[width=0.65\textwidth]{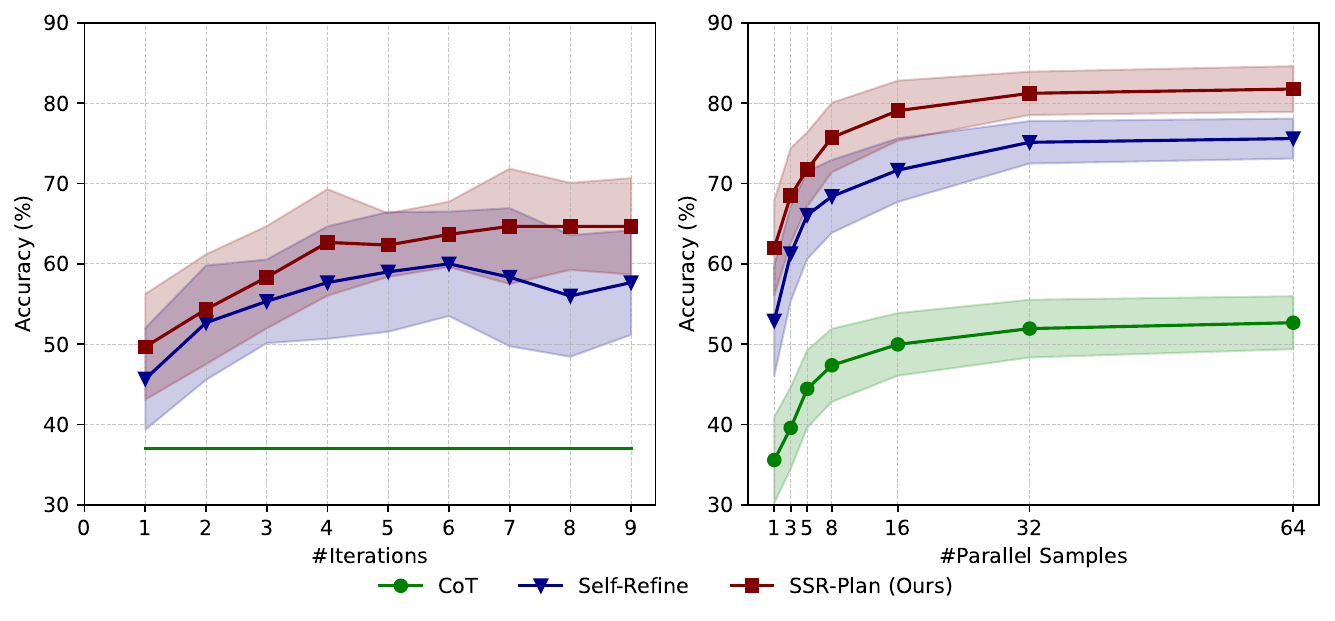}
    \vspace{-1.8em}
    \caption{
        \textbf{Performance of Sequential (Left) and Parallel (Right) Test-Time Scaling,} evaluated on AIME25~\citep{aime} with GPT-5-mini low-reasoning low-verbosity mode. 
    }
    \label{fig:tts-aime25-both}
\end{wrapfigure}
In our study, sequential scaling extends the number of iterations by $3\times$, with performance reported as Last-Round Accuracy (LR-Acc). Parallel scaling increases the number of parallel samples to 64, also reporting aggregated LR-Acc. Experiments are conducted on AIME25 with the \gptfive backbone (low-reasoning, low-verbosity). As baselines, we include basic CoT and Self-Refine. For Self-Refine and \ours, we perform an additional self-evaluation on the final reasoning trace and use the resulting 0-5 score for weighted best-of-$N$ (WBoN). For CoT, we apply majority voting (Maj@$N$). 

\textbf{The results are shown in \Figref{fig:tts-aime25-both}.}
\emph{On the sequential scaling side (left),} \ours consistently outperforms Self-Refine across all iteration counts. Accuracy improves steadily as the number of refinement iterations increases, with \ours showing stronger gains and greater stability than Self-Refine. In contrast, Self-Refine benefits from additional iterations but plateaus at a lower accuracy, confirming that iterative refinement is essential for improvement.
\emph{On the parallel scaling side (right),} all methods improve as the number of parallel samples increases, but \ours maintains a clear margin over Self-Refine and CoT. Notably, \ours reaches higher accuracy levels more quickly, suggesting that its Socratic step-level verification yields more consistent refinements, which aggregate effectively under parallel sampling. Self-Refine shows moderate improvements with larger sample sizes, while CoT lags behind, highlighting the importance of structured refinement.

\subsection{{Analysis: Granularity of Socratic Steps in \ours}}
\label{sec:experiments-granularity}
\begin{wrapfigure}{r}{0.65\textwidth}
    \vspace{-0.8em}
    \centering
    \includegraphics[width=0.65\textwidth]{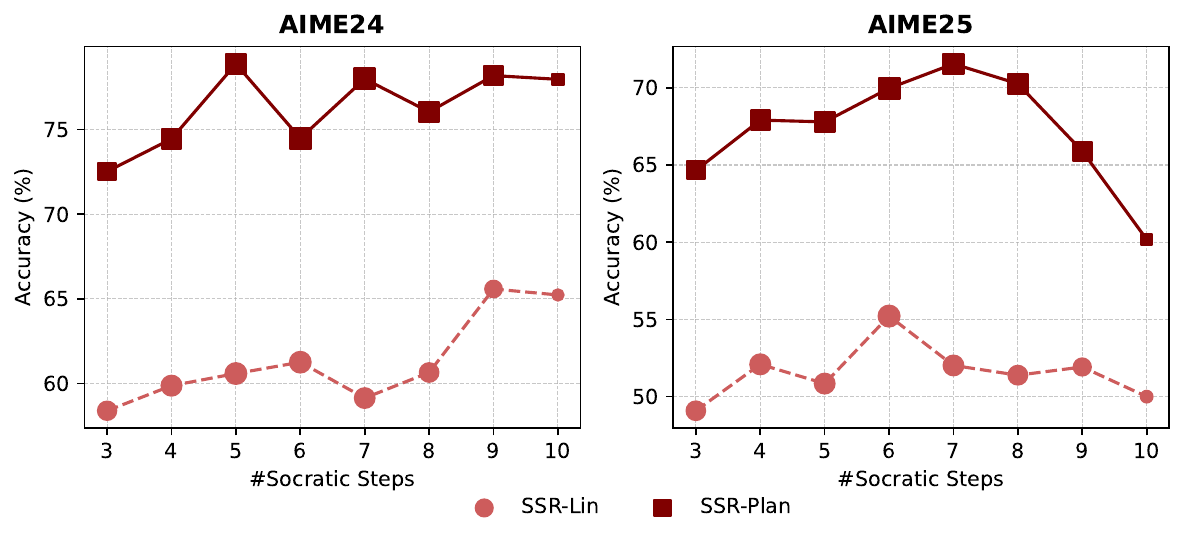}
    \vspace{-1.8em}
    \caption{
        \textbf{Performance of our \ours, with explicit control of granularity,} evaluated on AIME24 and AIME25~\citep{aime} with GPT-5-mini low-reasoning low-verbosity mode. 
        The marker size of each data point is proportional to the support size.
    }
    \label{fig:granularity}
\end{wrapfigure}
In this section, we investigate the effect of explicitly controlling decomposition granularity by varying the maximum number of Socratic steps. This is implemented by modifying the decomposition prompt: instead of instructing \emph{``Break down the reasoning process into a series of sub-questions,''} we use \emph{``Identify the most important milestones of the reasoning process and break it down into a series of sub-questions, with the number of sub-questions less than or equal to \{max\_steps\}.''} We then report iteration accuracy as a function of the actual number of Socratic steps produced by decomposition.
To isolate the effect of \ours, our main analysis is conducted with the Linear variant~(\ours-Lin), without adaptive gating or plan refinement, while also including \ours-Plan for reference (retaining steps that undergo Socratic decomposition). Note that setting the maximum number of steps to 1 reduces \ours to a final-answer evaluation via majority voting. Accordingly, we focus on the range of 3-10 steps in our experiments.

{\textbf{The results are reported in \Figref{fig:granularity}.}
For \ours-Lin, performance is relatively low and fluctuates with the number of Socratic steps, though a slight upward trend can be observed at higher step counts (e.g., 9-10 on AIME24). This suggests that finer-grained decomposition can sometimes help, but the effect is weak and unstable when refinement is applied without planning.
In contrast, the Plan-refinement variant (\ours-Plan) consistently achieves higher accuracy across all settings (possibly due to the gating mechanism of Self-Refine) and remains stable under varying levels of granularity. On AIME24, performance remains strong regardless of step count, while on AIME25, accuracy peaks around 6-7 steps and only drops when the decomposition becomes overly fine (10 steps). These results highlight that high-level plan refinement not only boosts overall accuracy but also makes \ours less sensitive to the specific choice of granularity, ensuring more reliable gains.
}

\section{Conclusion}
In this paper, we introduced Socratic Self-Refine (\ours), a novel iterative refinement framework that leverages step-level Socratic decomposition to evaluate and improve LLM reasoning. By explicitly modeling reasoning as a sequence of sub-questions and sub-answers, \ours provides fine-grained confidence estimation and enables targeted refinements where errors are most likely to occur. Across both mathematical and logical reasoning benchmarks, \ours consistently outperforms existing iterative refinement baselines, with the plan-refinement variant achieving the most robust gains. 
Beyond empirical performance, \ours highlights the importance of moving from outcome-level to process-level evaluation. By treating reasoning as a verifiable sequence of interpretable steps, our framework makes LLM outputs more transparent and opens the door to interventions that are more systematic than ad hoc self-correction.
We believe our \ours offers a valuable mechanism for controlling the reasoning trajectory, mitigating biases, and aligning model behavior more closely with human expectations.

\textbf{Limitations.}\quad 
Despite its advantages, \ours has several limitations. First, the step-level Socratic decomposition relies on LLM prompting, which can introduce noise and inconsistencies, particularly for problems with ambiguous or ill-posed sub-questions. Second, the computational cost of fine-grained verification is substantially higher than that of standard iterative refinement, limiting scalability to large datasets or long reasoning chains. Finally, our evaluation focuses primarily on mathematical and logical reasoning tasks; the generalizability of \ours to open-ended domains such as commonsense or multi-modal reasoning remains to be validated. 

\textbf{Future Work.}\quad
In future work, we aim to extend \ours to more diverse reasoning domains, including scientific and multimodal tasks, and explore tighter integration with training-time objectives. Another promising direction is developing more efficient confidence estimation to further reduce cost, as well as investigating human-in-the-loop settings where \ours can enhance interpretability and reliability.

\bibliography{ref}
\bibliographystyle{iclr2026_conference}

\appendix
\clearpage
\section*{\LARGE Appendix}

In \appref{app:llm-disclosure}, we describe the role of LLMs in our work.
In \appref{app:algorithm}, we present the full algorithmic description of our proposed \ours.
In \appref{app:implementation}, we provide our implementation details of the experiments, including: 
\begin{itemize}[nosep]
    \item \textbf{dataset details}~(\appref{app:datasets}),
    \item \textbf{baseline details}~(\appref{app:baselines}),
    \item \textbf{prompt templates} used in LLM reasoning~(\appref{app:prompts}),
\end{itemize}
Finally, in \appref{app:experiments}, we present additional experimental results, including:
\begin{itemize}[nosep]
    \item additional results on \textbf{\geminiflash}~(\appref{app:gemini}),
    \item additional results on \textbf{\emph{sequential} and \emph{parallel} test-time scaling}~(\appref{app:tts}),
    \item {a \textbf{breakdown detailed result} of \ours on Humanity's Last Exam~(HLE)~(\appref{app:hle}),}
    \item additional results on \textbf{\ours-as-a-Judge}~(\appref{app:judge}), 
    \item and \textbf{a qualitative analysis} on our \ours refinement~(\appref{app:qualitative}).
\end{itemize}

\section{LLM Usage Disclosure}
\label{app:llm-disclosure}
Large language models (LLMs) were used exclusively to help polish the writing of this paper by improving grammar, clarity, and readability. They did not contribute to research ideation, experimental design, data analysis, or the generation of scientific content. All technical contributions, claims, and conclusions are solely those of the authors.

\section{Algorithm}
\label{app:algorithm}
\begin{algorithm}[H]
\caption{\textbf{S}ocratic \textbf{S}elf-\textbf{R}efine~(\textbf{\ours})}\label{alg:ssr}
\begin{algorithmic}[1]
\INPUT
$\{\vx, \vx_{\text{dec}}, \vx_{\text{conf}}, \vx_{\text{ref}}\}$: prompt for original query, reasoning decomposition, confidence estimation, and refinement;\par
$\pi_{\vtheta}$: LLM policy;\par
$(\vz^{(0)}, \vy^{(0)})$: initial CoT reasoning and answer;\par
$K$: maximum refinement rounds;\par
$M$: number of parallel solves per sub-question for confidence;\par
$C_{\text{max}}$: maximum value of the confidence.\par

\STATE \textbf{(Optional)} $\{\vq_t\}_{t\in[T]}\sim \pi_\vtheta(\cdot\mid\vx, \vy^{(0)}, \vz^{(0)}, \vx_{\text{dec}})$.
Prompt $\pi_{\vtheta}$ to judge plan adequacy. If inadequate, refine once and update $(\vz^{(0)}, \vy^{(0)})$. \hfill$\rhd$ Optional plan check~(\Eqref{eq:plan}).
\FOR{$k = 1,\ldots,K$}
    \STATE $(\vz^{(k+1)}, \vy^{(k+1)}, C^{(k)})\leftarrow \operatorname{Self-Refine}(\vz^{(k)}, \vy^{(k)}).$ \hfill$\rhd$ Self-Refine Gating.
    \IF{$C^{(k)} = C_{\text{max}}$}
    %     \STATE  $(\vz^{(k+1)}, \vy^{(k+1)})\leftarrow \operatorname{Self-Refine}(\vz^{(k)}, \vy^{(k)}).$
    % \ELSE
        \STATE $\mS_T = \{(\vq_t,\va_t)\}_{t\in[T]} \sim \pi_{\vtheta}(\cdot \mid \vx,\vy^{(k)},\vz^{(k)},\vx_{\text{dec}})$. \hfill$\rhd$ \ours Decomposition (\Eqref{eq:core}).
        \FOR{$t = 1$ to $T$ in parallel}
            \STATE $\hat{\mA}_t = \{\hat{\va}_{ti}\}_{i\in[M]},\ \hat{\va}_{ti} \sim \pi_{\vtheta}(\cdot \mid \vq_t,\{\vs_i\}_{i<t},\vx)$. \hfill$\rhd$ Reference Set Sampling.
            \STATE $c_t \sim \pi_{\vtheta}(\cdot \mid \va_t,\hat{\mA}_t,\vx_{\text{conf}})$. \hfill$\rhd$ Confidence Estimation (\Eqref{eq:conf}).
        \ENDFOR
        \STATE $t' \leftarrow \arg\min_{t\in[T]} c_t$. \hfill$\rhd$ Pick weakest step
        \STATE $\va^{*}_{t'} \leftarrow \operatorname{maj\_vote}(\hat{\mA}_{t'})$. \hfill$\rhd$ Majority vote sub-answer
        \STATE $(\vz^{(k+1)}, \vy^{(k+1)}) \sim \pi_{\vtheta}(\cdot \mid \vx,\vy^{(k)},\vz^{(k)}, \vq^{(k)}_{t'},\va^{(k)}_{t'},\va^{*(k)}_{t'},\vx_{\text{ref}})$. \hfill$\rhd$ Round-$k$ \ours.
    \ENDIF
\ENDFOR
\OUTPUT
$(\vz^{(K)}, \vy^{(K)})$: refined reasoning and answer.
\end{algorithmic}
\end{algorithm}

\section{Implementation Details}
\label{app:implementation}
\appref{app:datasets} introduces the basic charactaristics of the adopted datasets; 
\appref{app:baselines} introduces the implementation details of the state-of-the-art iterative refinement baselines and our \ours. 
\appref{app:prompts} lists the prompt template we use for different methods. 

\subsection{Datasets}
\label{app:datasets}
\Tabref{tab:dataset_stats} shows the statistics of datasets in our experiments. These datasets span two different types of reasoning tasks and different difficulty levels, from moderate to highly challenging, covering both grade-school-level numerical reasoning and advanced symbolic mathematical tasks. This diversity in problem domains and difficulty ensures a comprehensive and representative assessment of the model's capabilities across varied reasoning scenarios.

\begin{table}[ht]
\begin{center}
\caption{Dataset Statistics.}
\vspace{-1em}
\label{tab:dataset_stats}
\resizebox{1\linewidth}{!}{
\setlength{\tabcolsep}{3pt}
\begin{tabular}{lccccc}
\toprule
\textbf{Dataset} & \textbf{\#Examples} & \textbf{Split} & \textbf{Task Type}  & \textbf{Language} &\textbf{Level}\\
\midrule
MATH-Level-5~\citep{hendrycks2021measuring} & 681 & Numerical-Answer Test Subset & Mathematical  & English & Moderate \\
AIME24~\citep{aime} & 30 & Full Set & Mathematical & English & Highly Challenging \\
AIME25~\citep{aime} & 30 & Full Set & Mathematical & English & Highly Challenging\\
{HLE}~\citep{phan2025humanity} & {915} & {Text-Only Math Subset} &{Mathematical} & {English} & {Extremely Challenging} \\
Zebra-Puzzle~\citep{stojanovski2025reasoning} & 100 & Randomly Synthesized & Logical & English & Moderate \\
Mini-Sudoku~\citep{stojanovski2025reasoning} & 100 & Randomly Synthesized & Logical & English & Moderate \\
\bottomrule
\end{tabular}
}
\end{center}
\end{table}

\subsection{Baselines and Our \ours}
\label{app:baselines}
We compare our proposed \textbf{S}ocratic \textbf{S}elf-\textbf{R}efine~(\ours) against several state-of-the-art iterative refinement reasoning frameworks. The detailed prompt templates are provided in the next section.
\begin{itemize}[nosep,leftmargin=24pt]
    \item \textbf{Self-Refine}~\citep{madaan2023self}: We follow the prompt template defined in LLM-as-a-Judge~\citep{zhou2025variation}, which produces feedback and scores for the model’s own response; the feedback is then used for refinement. We perform three refinement iterations, with each iteration independent of previous ones for conciseness.
    \item {\textbf{Debate}~\citep{du2023improving}: We adopt the official LLM-Debate code with two modifications: (i) using the unified CoT prompt for initial thought generation, as in this paper, and (ii) explicitly instructing each agent to \textbf{refine} its response based on the peer agent’s response. We run two agents for three iterations of debate, and for fair comparison, randomly select one of the final-round answers as the output.}
    \item \textbf{Monte Carlo Tree Self-Refine (MCTSr)}~\citep{zhang2024accessing}: We adopt the released code for reproducibility. Since the original prompt was designed for smaller open-source LLMs~\citep{touvron2023llama2,dubey2024llama} with format mismatches to our setting, we adapt the template while retaining the same verification prompt (as Self-Refine) and faithfully preserving the Monte Carlo Tree construction and exploration. The maximum number of iterations is set to four, following the original paper.
    \item \textbf{Atom-of-Thoughts (AoT)}~\citep{teng2025atom}: We mainly follow the released implementation. However, as the original decomposition restricts intermediate answers to purely numerical forms, which is limiting for challenging mathematical and logical reasoning, we slightly relax this constraint. For fair comparison, we set the maximum number of atoms to three, omit the final ``Ensemble'' step, and report only the last-iteration performance in \Tabref{tab:main}. Results with the ensemble step are reported separately in Column ``BoK-Acc'' of \Tabref{tab:main-pass-at-k}.
    \item \textbf{Forest-of-Thought (FoT)}~\citep{bi2024forest}: As a parallel scaling variant of MCTSr (ignoring early stopping), FoT is not directly evaluated. Nevertheless, MCTSr’s results in the “LR-Maj@5” column can be treated as an approximate proxy for FoT performance with tree size 5 and majority voting aggregation.
    \item \textbf{Linear \ours~(\ours-Lin, Ours)}: Each iteration proceeds as follows: (i) decompose the given CoT into Socratic steps; (ii) re-answer each sub-question multiple times, assuming prior steps are correct; (iii) identify the step with the lowest confidence score and refine based on the majority-voted sub-answer. We set the number of iterations to three for fairness.
    \item \textbf{Adaptive \ours~(\ours-Ada, Ours)}: At the beginning of each round, \ours-Ada first applies Self-Refine. If unreliable steps are identified with non-perfect scores, refinement proceeds via this efficient route. Otherwise (if Self-Refine fails or is overconfident), the method falls back to the full Socratic refinement.
    \item \textbf{\ours with Plan Refinement (\ours-Plan, Ours)}: Extends \ours-Ada by adding a preliminary plan refinement stage before iterative refinement.
\end{itemize}

\textbf{Shared LLM Configuration.}\quad
For \gptnano, we set the maximum token length to 16,384 and temperature to 0.6.
For \gptfive, we set the maximum completion length to 16,384 and temperature to 1.0.
For \geminiflash and {\geminilite}, we set the maximum completion length to 32,768 and temperature to 0.6.

\subsection{Prompt Templates}
\label{app:prompts}
This subsection presents the prompt templates used for the baselines and our \ours. The templates are identical for both mathematical and logical reasoning, except for a role specification: ``you are a precise math problem solver'' versus ``you are a precise logical reasoning problem solver.''
\begin{itemize}[nosep,leftmargin=24pt]
    \item \textbf{CoT:} uses \emph{Chain-of-Thought};
    \item \textbf{Self-Refine:} uses \emph{Verification} and \emph{Refine (Normal)};
    \item \textbf{MCTSr:} uses \emph{Verification} and \emph{Refine (Normal)};
    \item \textbf{AoT:} uses \emph{Decompose (AoT)}, \emph{Contract (AoT)}, and \emph{Ensemble};
    \item \textbf{\ours:} uses \emph{Decompose (\ours, Ours)}, \emph{Solve Sub-Question (\ours, Ours)}, \emph{Confidence Estimate (\ours, Ours)}, \emph{Reflection (\ours, Ours)}, and \emph{Refine (\ours, Ours)}. 
\end{itemize}

\begin{tcolorbox}[
    breakable,
    enhanced,
    left=-1cm, right=-1cm, top=2pt, bottom=2pt,
    enlarge top by=0.1cm, 
    enlarge bottom by=0.1cm, 
    title={\hspace{1cm} Chain-of-Thought}, 
    fonttitle=\bfseries\small
]
\begin{quote}
You are a precise math problem solver. Solve the given math problem step by step:\\

QUESTION: \{question\}\\

Please extend your chain of thought as much as possible; the longer the chain of thought, the better.\\

You can freely reason in your response, but please enclose the final answer within <answer></answer> tags (pure number without units and explanations).
\end{quote}
\end{tcolorbox}

\begin{tcolorbox}[
    breakable,
    enhanced,
    left=-1cm, right=-1cm, top=2pt, bottom=2pt,
    enlarge top by=0.1cm, 
    enlarge bottom by=0.1cm, 
    title={\hspace{1cm} Verification}, 
    fonttitle=\bfseries\small
]
\begin{quote}
Please act as an impartial judge and evaluate the correctness of the response provided by an AI assistant to the user prompt displayed below. You will be given the assistant's response.\\

When evaluating the assistant's response, identify any mistakes or inaccurate information. Be as objective as possible. Avoid any biases, such as order of responses, length, or stylistic elements like formatting.\\

Before providing an your final verdict, think through the judging process and output your thoughts as an explanation.\\

After providing your explanation, you must output a score of scale 0 to 5, where 0 represents you are completely certain that the response is incorrect and 5 represents you are completelycertain that the response is correct. Please enclose your score in <answer> and </answer> tags.

<|User Prompt|>\\
\{question\}\\
<|The Start of Assistant's Answer|>\\
\{response\}\\
<|The End of Assistant's Answer|>
\end{quote}
\end{tcolorbox}

\begin{tcolorbox}[
    breakable,
    enhanced,
    left=-1cm, right=-1cm, top=2pt, bottom=2pt,
    enlarge top by=0.1cm, 
    enlarge bottom by=0.1cm, 
    title={\hspace{1cm} Refine (Normal)}, 
    fonttitle=\bfseries\small
]
\begin{quote}
You are a precise math problem solver. Refine the provided solution to the given math problem, step-by-step, by meticulously addressing the judge's feedback (whose score is enclosed within <answer></answer> tags).\\

QUESTION: \{question\}\\
ORIGINAL SOLUTION: \{original\_cot\_response\}\\
JUDGE RESPONSE: \{judge\_response\}\\

Your task is to re-evaluate the original reasoning, identify where it went wrong based on the judge's comments, which should be enclosed within <evaluation></evaluation> tags; after that, construct a new, corrected chain of thought. Explain each step thoroughly. The more detailed and explicit your reasoning, the better.\\

You can freely reason in your response, but please enclose the final, numerical answer within <answer></answer> tags (pure number only, without units or explanations).
\end{quote}
\end{tcolorbox}

\begin{tcolorbox}[
    breakable,
    enhanced,
    left=-1cm, right=-1cm, top=2pt, bottom=2pt,
    enlarge top by=0.1cm, 
    enlarge bottom by=0.1cm, 
    title={\hspace{1cm} Decompose (AoT)}, 
    fonttitle=\bfseries\small
]
\begin{quote}
You are tasked with breaking down a math problem's reasoning process into a series of sub-questions.\\

Original Question: \{question\}\\
Complete Reasoning Process: \{trajectory\}\\

Instructions:
\begin{itemize}[nosep,leftmargin=24pt]
    \item Break down the reasoning process into a series of sub-questions.
    \item Each sub-question should:
    \begin{itemize}
        \item Be written in a clear, interrogative form.
        \item Be precise, unambiguous, and directly answerable from the provided reasoning or prior sub-question answers.
        \item Have a clear, **exact expression** as its answer (e.g., use fractions like `1/3`, symbolic representations like `pi`, or precise numerical values such as `1.0`). **Crucially, avoid approximations or rounding** unless the original question explicitly requires it.
        \item List the 0-based indexes of other sub-questions it depends on. This list can be empty if no prior sub-question answers are needed.
    \end{itemize}
    \item Dependencies are defined as information necessary to answer the current sub-question that:
    \begin{itemize}
        \item Does NOT come directly from the original question.
        \item MUST come from the answers of previous sub-questions.
    \end{itemize}
    \item **Stop generating sub-questions once the final answer to the Original Question has been fully derived from the reasoning process.** Do not include any subsequent or irrelevant steps that do not directly contribute to reaching the final answer.
\end{itemize}
Format your response as the following JSON object:
\small
\begin{verbatim}
{{
    "sub-questions": [
        {{
            "description": "<clear, precise interrogative question>",
            "answer": <exact expression of the answer>,
            "depend": [<indices of prerequisite sub-questions>]
        }},
        ...
    ],
    "answer": {answer}
}}
\end{verbatim}
\end{quote}
\end{tcolorbox}

\begin{tcolorbox}[
    breakable,
    enhanced,
    left=-1cm, right=-1cm, top=2pt, bottom=2pt,
    enlarge top by=0.1cm, 
    enlarge bottom by=0.1cm, 
    title={\hspace{1cm} Contract (AoT)}, 
    fonttitle=\bfseries\small
]
\begin{quote}
You are a math problem solver specializing in optimizing step-by-step reasoning processes. Your task is to optimize the existing reasoning trajectory into a more efficient, single self-contained question.\\
        
For the original question: \{question\}\\

Here are step-by-step reasoning process:\\
\{response\}\\

\{sub\_questions\}

Here are explanations of key concepts:
\begin{itemize}
    \item self-contained: The optimized question must be solvable independently, without relying on any external information
    \item efficient: The optimized question must be simpler than the original, requiring fewer reasoning steps (these steps are reduced because some solved independent sub-problems become known conditions in the optimized question or are excluded as incorrect explorations)
\end{itemize}
You can freely reason in your response, but please enclose the your optimized question within <question></question> tags.
\end{quote}
\end{tcolorbox}

\begin{tcolorbox}[
    breakable,
    enhanced,
    left=-1cm, right=-1cm, top=2pt, bottom=2pt,
    enlarge top by=0.1cm, 
    enlarge bottom by=0.1cm, 
    title={\hspace{1cm} Decompose (\ours, Ours)}, 
    fonttitle=\bfseries\small
]
\begin{quote}
You are tasked with breaking down a math problem's reasoning process into a series of **atomic** sub-questions.\\

Original Question: \{question\}\\
Complete Reasoning Process: \{trajectory\}\\

Instructions:
\begin{itemize}[nosep,leftmargin=24pt]
    \item Break down the reasoning process into a series of sub-questions.
    \item Each sub-question should:
    \begin{itemize}
        \item Be written in a clear, interrogative form.
        \item Be precise, unambiguous, and directly answerable from the provided reasoning or prior sub-question answers.
        \item Have a clear, **exact expression** as its answer (e.g., use fractions like `1/3`, symbolic representations like `pi`, or precise numerical values such as `1.0`). **Crucially, avoid approximations or rounding** unless the original question explicitly requires it.
        \item List the 0-based indexes of other sub-questions it depends on. This list can be empty if no prior sub-question answers are needed.
    \end{itemize}
    \item **Stop generating sub-questions once the final answer to the Original Question has been fully derived from the reasoning process.** Do not include any subsequent or irrelevant steps that do not directly contribute to reaching the final answer.
    \item The sub-question, sub-answer pairs should perfectly represent the reasoning process of the solution.
\end{itemize}
Format your response as the following JSON object:
\small
\begin{verbatim}
{{
    "sub-questions": [
        {{
            "description": "<clear, precise interrogative question>",
            "answer": <exact expression of the answer>,
        }},
        ...
    ],
    "answer": {answer}
}}
\end{verbatim}
\end{quote}
\end{tcolorbox}

\begin{tcolorbox}[
    breakable,
    enhanced,
    left=-1cm, right=-1cm, top=2pt, bottom=2pt,
    enlarge top by=0.1cm, 
    enlarge bottom by=0.1cm, 
    title={\hspace{1cm} Solve Sub-Question (\ours, Ours)}, 
    fonttitle=\bfseries\small
]
\begin{quote}
You are a precise math problem solver. Given the original question and the series of sub-questions and their answers which perfectly represent the reasoning process of the solution, think step by step and answer the next sub-question. Do not extend the reasoning process beyond this sub-question and enclose the answer within <answer></answer> tags.\\

Original question:\\
\{question\}\\

The series of sub-questions and their answers:\\
\{socratic\_reasoning\_trajectory\}\\

The next sub-question to be answered:\\
\{next\_sub\_question\}
\end{quote}
\end{tcolorbox}

\begin{tcolorbox}[
    breakable,
    enhanced,
    left=-1cm, right=-1cm, top=2pt, bottom=2pt,
    enlarge top by=0.1cm, 
    enlarge bottom by=0.1cm, 
    title={\hspace{1cm} Confidence Estimate (\ours, Ours)}, 
    fonttitle=\bfseries\small
]
\begin{quote}
You are a math expert. Given the a math expression as the prediction and a list of reference answers, determine the confidence of the prediction.\\

The prediction is:\\
\{prediction\}\\

The reference answers are:\\
\{answers\}\\

Please answer with a number of scale 0 to 5 that represents the confidence of the prediction. 0 means the prediction does not match any of the reference answers. 5 means the prediction matches the reference answers perfectly. If you cannot determine the confidence, please answer with -1. Enclose the answer within <answer></answer> tags.
\end{quote}
\end{tcolorbox}

\begin{tcolorbox}[
    breakable,
    enhanced,
    left=-1cm, right=-1cm, top=2pt, bottom=2pt,
    enlarge top by=0.1cm, 
    enlarge bottom by=0.1cm, 
    title={\hspace{1cm} Reflection (\ours, Ours)}, 
    fonttitle=\bfseries\small
]
\begin{quote}
Wait, in the sub-step of "\{wrong\_question\}", the answer is "\{wrong\_answer\}", but after careful re-evaluating the process, I think that the actual answer to this sub-question should be "\{revised\_answer\}".
\end{quote}
\end{tcolorbox}

\begin{tcolorbox}[
    breakable,
    enhanced,
    left=-1cm, right=-1cm, top=2pt, bottom=2pt,
    enlarge top by=0.1cm, 
    enlarge bottom by=0.1cm, 
    title={\hspace{1cm} Refine (\ours, Ours)}, 
    fonttitle=\bfseries\small
]
\begin{quote}
\{cot\_instruction\}\\

\{cot\_reasoning\_trace\}\\

\{reflection\}\\

Let's re-evaluate the reasoning process based on your reflection. Enclose it within <evaluation></evaluation> tags. After that, let's reasoning step by step again to solve the original question. This time, you should address the specific issue identified in your own re-evaluation. Finally,enclose the final answer within <answer></answer> tags."
\end{quote}
\end{tcolorbox}

\begin{tcolorbox}[
    breakable,
    enhanced,
    left=-1cm, right=-1cm, top=2pt, bottom=2pt,
    enlarge top by=0.1cm, 
    enlarge bottom by=0.1cm, 
    title={\hspace{1cm} Ensemble}, 
    fonttitle=\bfseries\small
]
\begin{quote}
You are a precise math problem solver. Compare then synthesize the best answer from multiple solutions to solve the following question.\\

QUESTION: \{question\}\\

SOLUTIONS:\\
\{solutions\}\\

Please extend your chain of thought as much as possible; the longer the chain of thought, the better.\\

You can freely reason in your response, but please enclose the final answer within <answer></answer> tags (pure number without units and explanations).
\end{quote}
\end{tcolorbox}

\begin{tcolorbox}[
    breakable,
    enhanced,
    left=-1cm, right=-1cm, top=2pt, bottom=2pt,
    enlarge top by=0.1cm, 
    enlarge bottom by=0.1cm, 
    title={\hspace{1cm} {LLM-as-a-Judge for Humanity's Last Exam (HLE) Evaluation}}, 
    fonttitle=\bfseries\small
]
\begin{quote}
Judge whether the following [candidate\_answer] to [question] is correct or not based on the precise and unambiguous [correct\_answer] below.\\

[question]: \{question\} 

[correct\_answer]: \{correct\_answer\} 

[candidate\_answer]: \{candidate\_answer\} \\

Your judgement must be in the format and criteria specified below:\\

reasoning: Explain why the [candidate\_answer] is correct or incorrect based on [correct\_answer], focusing only on if there are meaningful differences between [correct\_answer] and the [candidate\_answer]. Do not comment on any background to the problem, do not attempt to solve the problem, do not argue for any answer different than [correct\_answer], focus only on whether the answers match.\\

correct: Answer '1' if [candidate\_answer] matches the [correct\_answer] given above, or is within a small margin of error for numerical problems. Answer '0' otherwise, i.e. if there if there is any inconsistency, ambiguity, non-equivalency, or if the extracted answer is incorrect.\\

Please enclose your reasoning within <reasoning></reasoning> tags, and your correct answer within <correct></correct> tags.
\end{quote}
\end{tcolorbox}

\section{Additional Experimental Results}
\label{app:experiments}
\appref{app:gemini} reports additional results on a strong model, \geminiflash.
\appref{app:tts} provides further experiments on both \emph{sequential} and \emph{parallel} test-time scaling.
\appref{app:judge} presents results using \ours as an LLM judge, offering deeper insights into its underlying mechanism.
Finally, \appref{app:qualitative} includes qualitative examples that illustrate the behavior of \ours in practice.

\subsection{Additional Results of \geminiflash}
\label{app:gemini}

\begin{table*}[h]
\caption{
    \textbf{Performance of Iterative Refinement-Based Reasoning Methods.} 
    \textbf{LR-Acc:} Last-round refinement's accuracy, yielded by 10 repeated experiments;
    \textbf{Pass@K:} Pass-at-K refinements' accuracy (at lease one of K iterations gets the answer correct).
    \textbf{LR-Maj@5:} Last-round refinement's accuracy of majority voting with 5 samples in parallel, yielded by 50 repeated experiments.
    \textbf{Boldface} and \underline{underlining} denote the best and the second-best performance, respectively. 
}
\vspace{-1.2em}
\begin{center}
\resizebox{1\linewidth}{!}{%
\setlength{\tabcolsep}{3pt}
\begin{tabular}{l ccc ccc ccc}
	\toprule[0.12em] 
    \multirow{2}{*}[-0.25em]{\textbf{Method}} 
    & \multicolumn{3}{c}{\textbf{AIME24}} 
    & \multicolumn{3}{c}{\textbf{AIME25}} 
    & \multicolumn{3}{c}{\textbf{Zebra-Puzzle}}
    \\
    \cmidrule(lr){2-4} \cmidrule(lr){5-7} \cmidrule(lr){8-10} 
     
    & \emph{LR-Acc} & \emph{Pass@K} & \emph{LR-Maj@5} 
    & \emph{LR-Acc} & \emph{Pass@K} & \emph{LR-Maj@5} 
    & \emph{LR-Acc} & \emph{Pass@K} & \emph{LR-Maj@5} 
    \\

    \midrule

    \multicolumn{10}{c}{{\geminilite}} 
    \\
    \midrule

    CoT
    & 59.00\scriptsize{$\pm$4.48}
    & -
    & 68.53\scriptsize{$\pm$3.14}
    & 44.85\scriptsize{$\pm$5.92}
    & -
    & 52.47\scriptsize{$\pm$4.51}
    & 74.00\scriptsize{$\pm$2.72}
    & -
    & 84.18\scriptsize{$\pm$1.77}
    \\

    Self-Refine
    & 60.67\scriptsize{$\pm$6.29}
    & 66.33\scriptsize{$\pm$3.14}
    & 71.07\scriptsize{$\pm$2.44}
    & 50.00\scriptsize{$\pm$4.22}
    & 55.33\scriptsize{$\pm$3.40}
    & 61.20\scriptsize{$\pm$5.11}
    & 76.20\scriptsize{$\pm$3.46}
    & 82.00\scriptsize{$\pm$1.95}
    & 87.30\scriptsize{$\pm$1.50}
    \\

    MCTSr %\scriptsize{(d=4)}
    & 63.00\scriptsize{$\pm$6.40}
    & -
    & 69.33\scriptsize{$\pm$3.27}
    & 49.00\scriptsize{$\pm$6.51}
    & -
    & 57.07\scriptsize{$\pm$5.11}
    & 78.50\scriptsize{$\pm$2.84}
    & -
    & 86.62\scriptsize{$\pm$1.38}
    \\

    AoT
    & 64.67\scriptsize{$\pm$4.27}
    & 71.33\scriptsize{$\pm$4.27}
    & 72.13\scriptsize{$\pm$2.38}
    & 46.67\scriptsize{$\pm$4.47}
    & 51.00\scriptsize{$\pm$2.13}
    & 49.27\scriptsize{$\pm$3.35}
    & 62.40\scriptsize{$\pm$3.83}
    & 82.50\scriptsize{$\pm$1.96}
    & 79.18\scriptsize{$\pm$2.70}
    \\

    \rowcolor{lightergray}
    \ours-Lin~(Ours)
    & \underline{70.00\scriptsize{$\pm$4.47}}
    & \underline{72.00\scriptsize{$\pm$4.52}}
    & \underline{73.20\scriptsize{$\pm$2.00}}
    & \underline{55.33\scriptsize{$\pm$2.67}}
    & \underline{57.33\scriptsize{$\pm$4.16}}
    & \underline{60.93\scriptsize{$\pm$3.77}}
    & \textbf{82.60\scriptsize{$\pm$2.24}}
    & 85.70\scriptsize{$\pm$2.24}
    & 87.92\scriptsize{$\pm$1.65}
    \\

    \rowcolor{lightergray}
    \ours-Ada~(Ours)
    & 68.00\scriptsize{$\pm$5.21}
    & 70.67\scriptsize{$\pm$4.67}
    & 72.07\scriptsize{$\pm$1.87}
    & 54.00\scriptsize{$\pm$4.16}
    & \underline{57.33\scriptsize{$\pm$5.93}}
    & 59.07\scriptsize{$\pm$3.47}
    & \underline{82.40\scriptsize{$\pm$2.15}}
    & \underline{86.30\scriptsize{$\pm$2.10}}
    & \textbf{89.16\scriptsize{$\pm$1.71}}
    \\

    \rowcolor{lightergray}
    \ours-Plan~(Ours)
    & \textbf{70.33\scriptsize{$\pm$4.07}}
    & \textbf{73.33\scriptsize{$\pm$3.33}}
    & \textbf{73.87\scriptsize{$\pm$2.34}}
    & \textbf{56.67\scriptsize{$\pm$5.16}}
    & \textbf{61.00\scriptsize{$\pm$5.17}}
    & \textbf{65.47\scriptsize{$\pm$3.45}}
    & 81.10\scriptsize{$\pm$2.95}
    & \textbf{86.50\scriptsize{$\pm$2.33}}
    & \underline{87.48\scriptsize{$\pm$2.23}}
    \\
     
    \midrule

    \multicolumn{10}{c}{\geminiflash} 
    \\
    \midrule

    CoT
    & 81.85\scriptsize{$\pm$2.77}
    & -
    & 85.60\scriptsize{$\pm$1.55}
    & 68.00\scriptsize{$\pm$4.52}
    & -
    & 72.47\scriptsize{$\pm$3.99}
    & 67.44\scriptsize{$\pm$1.89}
    & -
    & 76.12\scriptsize{$\pm$1.92}
    \\

    Self-Refine
    & 82.96\scriptsize{$\pm$3.67}
    & 87.41\scriptsize{$\pm$3.05}
    & 88.87\scriptsize{$\pm$2.46}
    & 76.33\scriptsize{$\pm$7.06}
    & 81.00\scriptsize{$\pm$4.23}
    & 84.60\scriptsize{$\pm$2.48}
    & 75.25\scriptsize{$\pm$2.95}
    & 77.00\scriptsize{$\pm$3.32}
    & 88.98\scriptsize{$\pm$1.49}
    \\

    % {Debate}
    % \\

    MCTSr %\scriptsize{(d=4)}
    & 83.00\scriptsize{$\pm$4.07}
    & -
    & 86.67\scriptsize{$\pm$2.31}
    & 70.95\scriptsize{$\pm$7.50}
    & -
    & 77.73\scriptsize{$\pm$2.78}
    & 75.60\scriptsize{$\pm$2.94}
    & -
    & 85.68\scriptsize{$\pm$1.91}
    \\

    AoT
    & 81.67\scriptsize{$\pm$1.67}
    & 85.33\scriptsize{$\pm$2.21}
    & 86.13\scriptsize{$\pm$2.86}
    & 70.74\scriptsize{$\pm$5.62}
    & 75.19\scriptsize{$\pm$6.50}
    & 78.40\scriptsize{$\pm$2.60}
    & 54.71\scriptsize{$\pm$3.49}
    & 86.14\scriptsize{$\pm$1.88}
    & 65.74\scriptsize{$\pm$2.39}
    \\

    \rowcolor{lightergray}
    \ours-Lin~(Ours)
    & \textbf{86.30\scriptsize{$\pm$3.99}}
    & \textbf{90.37\scriptsize{$\pm$4.29}}
    & \textbf{90.93\scriptsize{$\pm$2.98}}
    & \textbf{79.26\scriptsize{$\pm$4.66}}
    & 83.33\scriptsize{$\pm$4.16}
    & \textbf{88.47\scriptsize{$\pm$3.14}}
    & \textbf{87.62\scriptsize{$\pm$2.18}}
    & \textbf{89.75\scriptsize{$\pm$2.54}}
    & \textbf{92.30\scriptsize{$\pm$1.36}}
    \\

    \rowcolor{lightergray}
    \ours-Ada~(Ours)
    & 82.50\scriptsize{$\pm$4.00}
    & 87.50\scriptsize{$\pm$3.23}
    & 88.33\scriptsize{$\pm$1.67}
    & 76.30\scriptsize{$\pm$6.37}
    & \textbf{84.44\scriptsize{$\pm$4.71}}
    & \underline{87.27\scriptsize{$\pm$2.72}}
    & \underline{87.14\scriptsize{$\pm$1.96}}
    & \underline{89.00\scriptsize{$\pm$1.69}}
    & 91.86\scriptsize{$\pm$1.30}
    \\

    \rowcolor{lightergray}
    \ours-Plan~(Ours)
    & \underline{84.17\scriptsize{$\pm$4.00}}
    & \underline{89.17\scriptsize{$\pm$3.63}}
    & \underline{89.67\scriptsize{$\pm$1.00}}
    & \underline{78.00\scriptsize{$\pm$6.00}}
    & \underline{84.00\scriptsize{$\pm$4.42}}
    & 86.73\scriptsize{$\pm$3.16}
    & 86.50\scriptsize{$\pm$2.69}
    & \underline{89.00\scriptsize{$\pm$2.50}}
    & \underline{92.06\scriptsize{$\pm$1.39}}
    \\
    
    \bottomrule[0.12em]
    \end{tabular}
}
\end{center}
\label{tab:main-gemini}
\vspace{-0em}
\end{table*}

\textbf{We further report results of applying \ours to a different model family, including a smaller and faster model, \geminilite, and a stronger model, \geminiflash}~\citep{comanici2025gemini}. Owing to its exceptionally strong mathematical and logical reasoning ability, two benchmarks used in the main body (MATH-Level-5 and Mini-Sudoku) are no longer suitable for differentiating framework performance, as naive CoT already solves nearly all questions correctly. Therefore, we report results only on the remaining three datasets, following the same evaluation protocols described in \Secref{sec:experiments}.

When applied to the stronger \geminiflash model, our \ours variants continue to demonstrate consistent improvements over baseline iterative refinement methods. On AIME24 and AIME25, \ours-Lin achieves the highest LR-Acc and LR-Maj@5, while \ours-Ada and \ours-Plan deliver highly competitive results, particularly in terms of Pass@K, reflecting their ability to exploit refinement opportunities even when the base model is already very strong. The gains are especially notable on AIME25, where \ours-Ada substantially outperforms all baselines in both LR-Acc and Pass@K, indicating the effectiveness of adaptively switching between efficient self-refinement and more costly Socratic refinement. On Zebra-Puzzle, all three variants of \ours surpass or match the best-performing baselines, with \ours-Lin again delivering the strongest overall results. These findings confirm that even for a state-of-the-art reasoning model like \geminiflash, our refinement strategies provide additional benefits, reinforcing their generality and scalability across model families and task types.

\subsection{Additional Results of Test-Time Scaling at Larger Scale}
\label{app:tts}
\begin{figure}[t] 
\centering 
\includegraphics[width=0.7\textwidth]{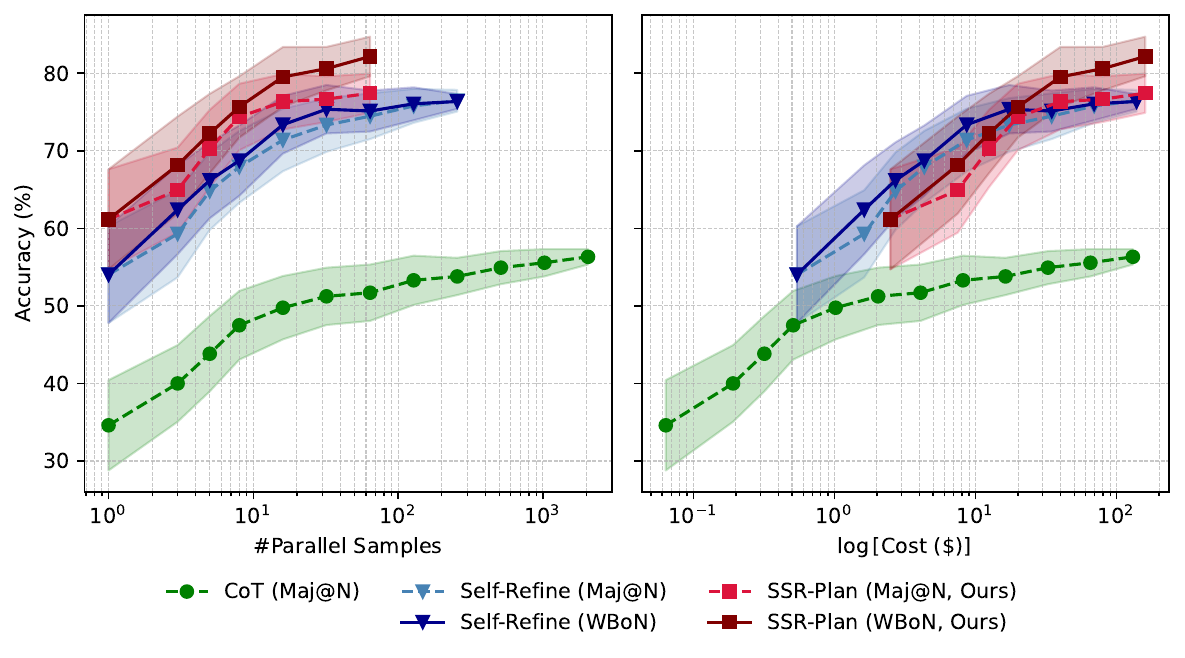}
\vspace{-1em}
\caption{
    \textbf{Performance of Parallel Test-Time Scaling,} evaluated on AIME25 with \gptfive low-reasoning low-verbosity mode. 
        Iterative refinement (both Self-Refine~\citep{madaan2023self} and our \ours) holds non-trivial advantage against CoT~\citep{wei2022chain} in terms of absolute performance and budget control. Our \ours outperforms the baselines under the same budget, with \ours's confidence estimation playing a crucial role. 
} 
\label{fig:tts-aime25-parallel} 
\end{figure}

\begin{figure}[t]
    \centering
    \includegraphics[width=0.7\textwidth]{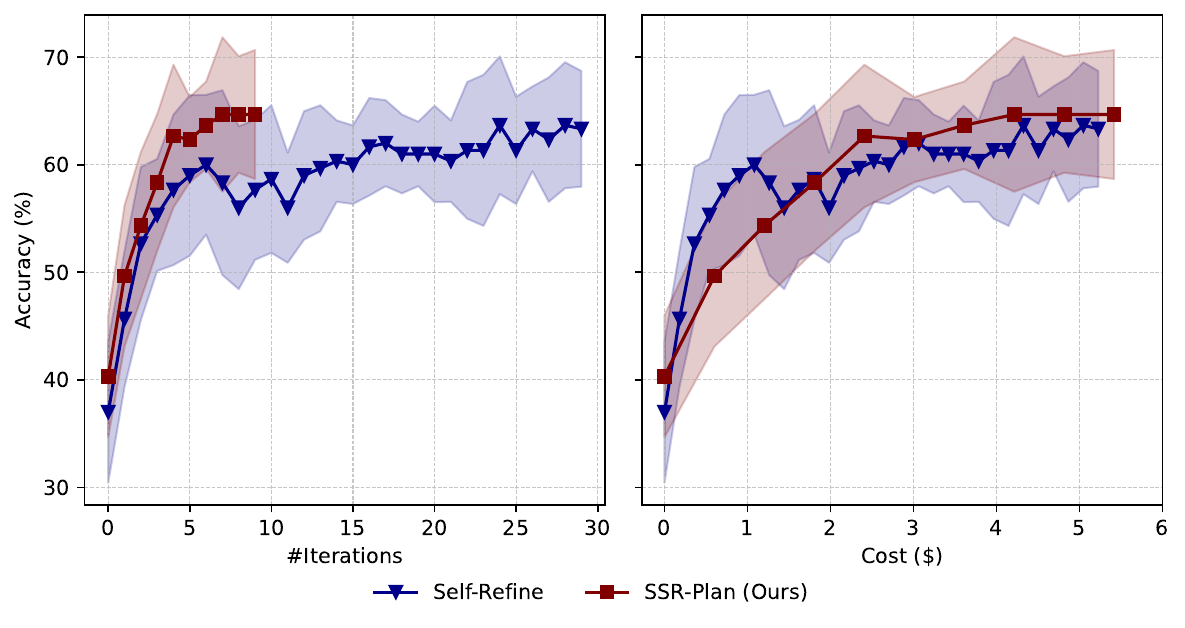}
    \vspace{-0.8em}
    \caption{
        \textbf{Performance of Iterative Test-Time Scaling,} evaluated on AIME25 with \gptfive low-reasoning low-verbosity mode. 
    }
    \label{fig:tts-aime25-linear}
\end{figure}

Applying iterative refinement, even for a single round, inevitably increases computation and latency at test time. Thus, comparisons restricted to a fixed number of iterations, as in \Secref{sec:experiments-tts}, may be unfair or incomplete. To more fairly assess efficiency, we examine the test-time scaling behavior of our \ours relative to baselines under comparable computational cost. \textbf{The results are presented in \Figref{fig:tts-aime25-parallel} (parallel scaling) and \Figref{fig:tts-aime25-linear} (sequential scaling).}

In the parallel scaling setting (\Figref{fig:tts-aime25-parallel}), both Self-Refine and our \ours substantially outperform vanilla CoT across all compute budgets, confirming that iterative refinement provides clear gains when additional samples are available. Importantly, our \ours consistently yields higher accuracy than Self-Refine under the same budget, demonstrating that confidence-aware step selection and plan refinement lead to more efficient use of compute.
In the sequential scaling setting (\Figref{fig:tts-aime25-linear}), a similar trend emerges: while performance plateaus quickly for Self-Refine, \ours continues to improve steadily with additional iterations, particularly in the early- to mid-cost regime. This suggests that \ours better leverages iterative opportunities, correcting errors that Self-Refine either overlooks or misjudges. Taken together, these results demonstrate that \ours not only provides stronger single-iteration performance but also scales more effectively under increased compute, striking a favorable balance between accuracy and cost.

\subsection{{Detailed Results of Humanity's Last Exam (HLE)}}
\label{app:hle}
{
\textbf{\Tabref{tab:hle-numerical} and \Tabref{tab:hle-non-numerical} present a detailed breakdown of \ours performance on the numerical and non-numerical subsets of Humanity’s Last Exam~(HLE)~\citep{phan2025humanity}.} On the numerical subset, \ours achieves substantial gains over both CoT and Self-Refine, improving accuracy by up to 8.89\% with \texttt{GPT-5-mini} and 5.23\% with the full \texttt{GPT-5}.
In contrast, on the non-numerical subset, improvements are smaller or even negative, particularly for \texttt{GPT-5}, where Self-Refine and \ours both slightly underperform CoT. This disparity suggests that non-numerical problems, often involving abstract or conceptual reasoning, may benefit less from explicit step-level self-verification and refinement, as it can introduce semantic drift or over-justification. Overall, these results demonstrate that \ours is especially effective for precise, calculation-heavy reasoning but may require further adaptation for more open-ended or qualitative tasks.
}

\begin{table}[t]
\caption{
Accuracies (\%) of iterative refinement-based reasoning methods on the 478-question challenging math subset (w/ numerical ground-truth answer) of Humanity’s Last Exam~(HLE)~\citep{phan2025humanity}, with \gptfive and \texttt{GPT-5} (medium reasoning, medium verbosity).
}
\vspace{-0.6em}
\centering
\resizebox{0.7\linewidth}{!}{%
\setlength{\tabcolsep}{14pt}
\begin{tabular}{lccc}
    \toprule[0.12em]
    \textbf{Model}
    & \textbf{CoT}
    & \textbf{Self-Refine}
    & \textbf{\ours-Plan~(Ours)}
    \\
    \midrule 

    \gptfive 
    & 17.78
    & 23.85~{\textcolor{ForestGreen}{(+6.07)}}
    & \textbf{26.57~{\textcolor{ForestGreen}{(+8.89)}}}
    \\
     
    \texttt{GPT-5}
    & 30.33
    & 33.89~{\textcolor{ForestGreen}{(+3.56)}}
    & \textbf{35.56~{\textcolor{ForestGreen}{(+5.23)}}}
    \\
     
    \bottomrule[0.12em]
    \end{tabular}
 }
% \end{center}
\label{tab:hle-numerical}
\vspace{0.8em}
\end{table}

\begin{table}[t]
\caption{
{
Accuracies (\%) of iterative refinement-based reasoning methods on the 437-question challenging math subset (w/ non-numerical ground-truth answer) of Humanity’s Last Exam~(HLE)~\citep{phan2025humanity}, with \gptfive and \texttt{GPT-5} (medium reasoning, medium verbosity).
}
}
\vspace{-0.6em}
\centering
\resizebox{0.7\linewidth}{!}{%
\setlength{\tabcolsep}{14pt}
\begin{tabular}{lccc}
    \toprule[0.12em]
    \textbf{Model}
    & \textbf{CoT}
    & \textbf{Self-Refine}
    & \textbf{\ours-Plan~(Ours)}
    \\
    \midrule 

    \gptfive 
    & 14.42
    & 12.81~{\textcolor{Red}{(-1.61)}}
    & \textbf{16.02~{\textcolor{ForestGreen}{(+1.60)}}}
    \\
     
    \texttt{GPT-5}
    & \textbf{25.40}
    & 18.08~{\textcolor{Red}{(-7.32)}}
    & 23.11~{\textcolor{Red}{(-2.29)}}
    \\
     
    \bottomrule[0.12em]
    \end{tabular}
 }
\label{tab:hle-non-numerical}
\vspace{1em}
\end{table}

\subsection{Additional Results of \ours-as-a-Judge}
\label{app:judge}
\begin{figure}[t] 
\centering 
\vspace{-0em}
\includegraphics[width=1\textwidth]{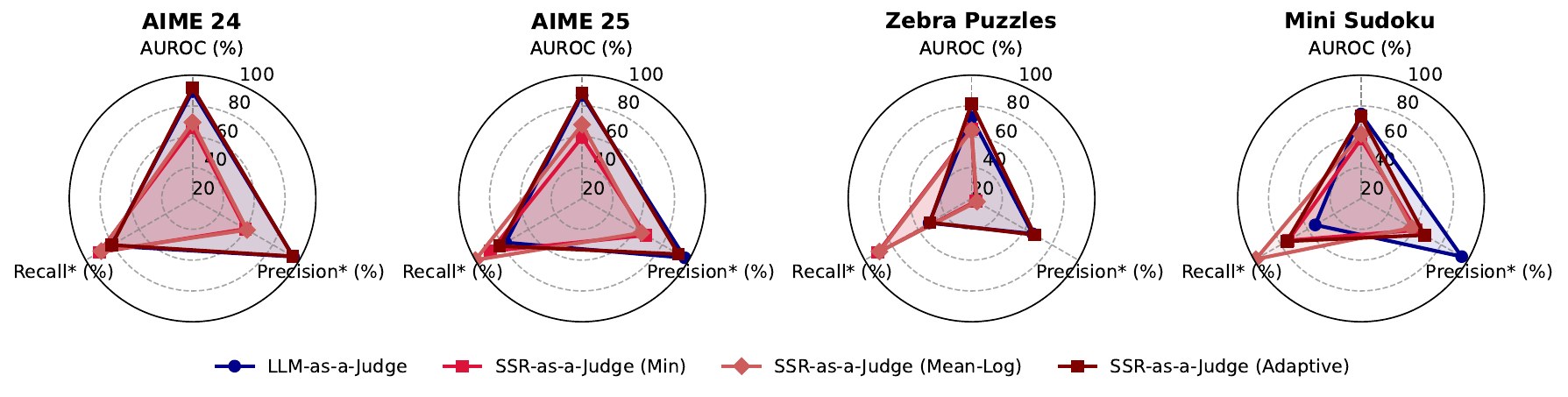}
\vspace{-2em}
\caption{
    \textbf{Self-Evaluation Performance of \ours-as-a-Judge and LLM-as-a-Judge,} evaluated with \gptfive.
}
\vspace{-0em}
\label{fig:judge}
\end{figure}
To better understand the strengths of \ours, we further assess its self-evaluation quality and compare it with the LLM-as-a-Judge framework~\citep{gu2024survey}.
We evaluate the self-evaluation ability on the four datasets we use in the main body, and we further include the results on ProcessBench~\citep{zhang2025lessons}. For self-evaluation, due to the smaller dataset sizes of AIME24 and AIME25, we sample 100 parallel reasoning traces per question, yielding datasets of 3,000 examples each. For logical reasoning, we sample 10 traces per question, resulting in datasets of 1,000 examples each. In the LLM-as-a-Judge setting, the model is prompted to provide both feedback and a confidence score on a 0-5 scale. For \ours, we perform a single iteration of Socratic step decomposition and confidence estimation of each step. All experiments run with \gptfive low-reasoning low-verbosity mode. 
Since \ours produces step-level confidence scores $\mC_T=\{c_t\}_{t\in[T]}$ for the Socratic steps $\mS_T=\{\vs_t\}_{t\in[T]}$, these must be aggregated into a single score to represent overall response quality. We show the result of (i)~\textbf{Min}~($\min\{c_t\}_{t\in [T]}$), the weakest step confidence; (ii)~\textbf{Mean-Log}~($\frac{1}{T}\sum_{t=1}^{T}\log c_t$), a length-normalized version inspired by confidence and uncertainty estimation in sequence modeling~\citep{zhang2025token}; and (iii)~\ours-Ada with \textbf{Mean}. 

We formulate the evaluation of a judge’s ability as a problem of incorrect reasoning trace detection, where incorrect responses are labeled as positives. We report three correlation-based metrics: Area Under the Receiver Operating Characteristic Curve~(\textbf{AUROC}), \textbf{Precision$^*$} and \textbf{Recall$^*$} at the optimal classification threshold~\citep{hanley1982meaning,boyd2013area,farquhar2024detecting,ye2025uncertainty,zhang2025token}, which together measure how well confidence scores distinguish between correct and incorrect responses.

\begin{figure}[t] 
\centering 
\vspace{0em}
\includegraphics[width=1\textwidth]{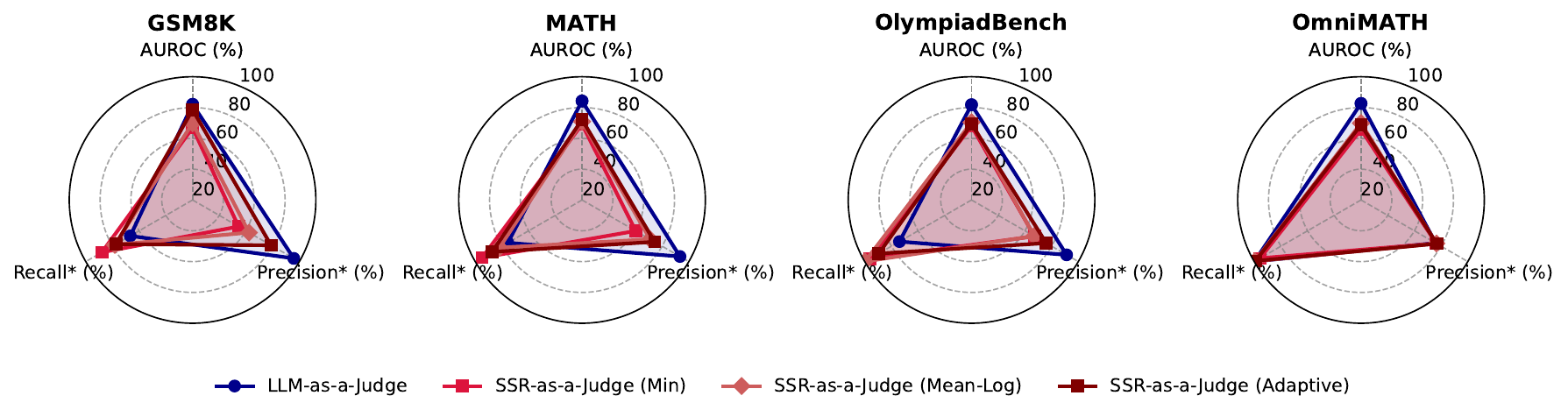}
\vspace{-1.5em}
\caption{
    \textbf{Performance of \ours-as-a-Judge and LLM-as-a-Judge,} evaluated on ProcessBench~\citep{zhang2025lessons}, with \gptnano.
}
\vspace{-0.0em}
\label{fig:process-bench}
\end{figure}

\textbf{The results are shown in \Figref{fig:judge} and \Figref{fig:process-bench}.} 
Somewhat unexpectedly, across most evaluation metrics, the judging ability of \ours does not surpass the basic LLM-as-a-Judge. This is evident in consistently lower AUROC, suggesting that the confidence scores produced by \ours contain more noise and thus yield less balanced evaluations.
\emph{Why, then, does \ours still outperform baselines as an iterative refinement framework?} As illustrated in \Figref{fig:judge}, the key lies in its complementary role to Self-Refine. 
While \ours lags behind LLM-as-a-Judge in AUROC,, it consistently achieves much higher recall of incorrect reasoning traces, particularly on logical reasoning benchmarks such as Zebra Puzzle and Mini-Sudoku.
This broader coverage allows \ours to catch errors that Self-Refine often misses, even if it introduces additional noise. The mechanism behind \ours-Ada can thus be understood as three cascading factors:
\begin{itemize}[nosep,leftmargin=24pt]
    \item \textbf{High precision of LLM-as-a-Judge:} when used in Self-Refine, it reliably identifies problematic reasoning traces, but often misses a large portion of incorrect ones.  
    \item \textbf{High coverage of \ours:} it captures and provides useful signals for truly problematic steps in reasoning, though at the cost of introducing some unreliable feedback for feedback.  
    \item \textbf{Inherent robustness of LLMs:} during refinement, LLMs can withstand noisy refinement feedback, serving as a safeguard that enables recovery and improvement despite occasional errors.  
\end{itemize}

\subsection{Qualitative Analysis}
\label{app:qualitative}
\begin{figure}[t] 
\centering
\vspace{0em}
\includegraphics[width=1\textwidth]{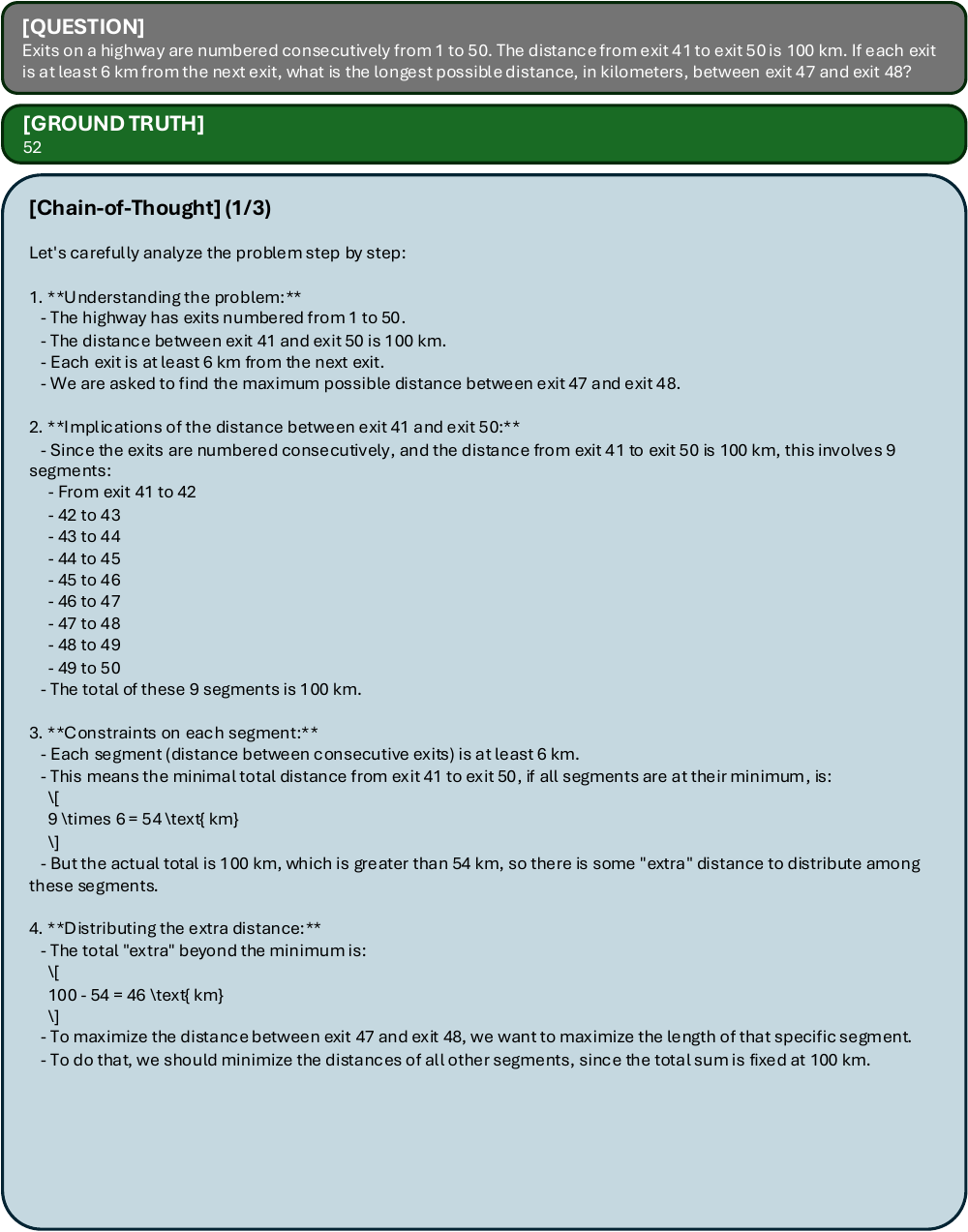}
\caption{
    \textbf{Qualitative Result, Concrete Example (1/7):} Chain-of-Thought Response (1/3).
}
\vspace{-0.0em}
\label{fig:qual-1}
\end{figure}

\begin{figure}[t] 
\centering 
\vspace{0em}
\includegraphics[width=1\textwidth]{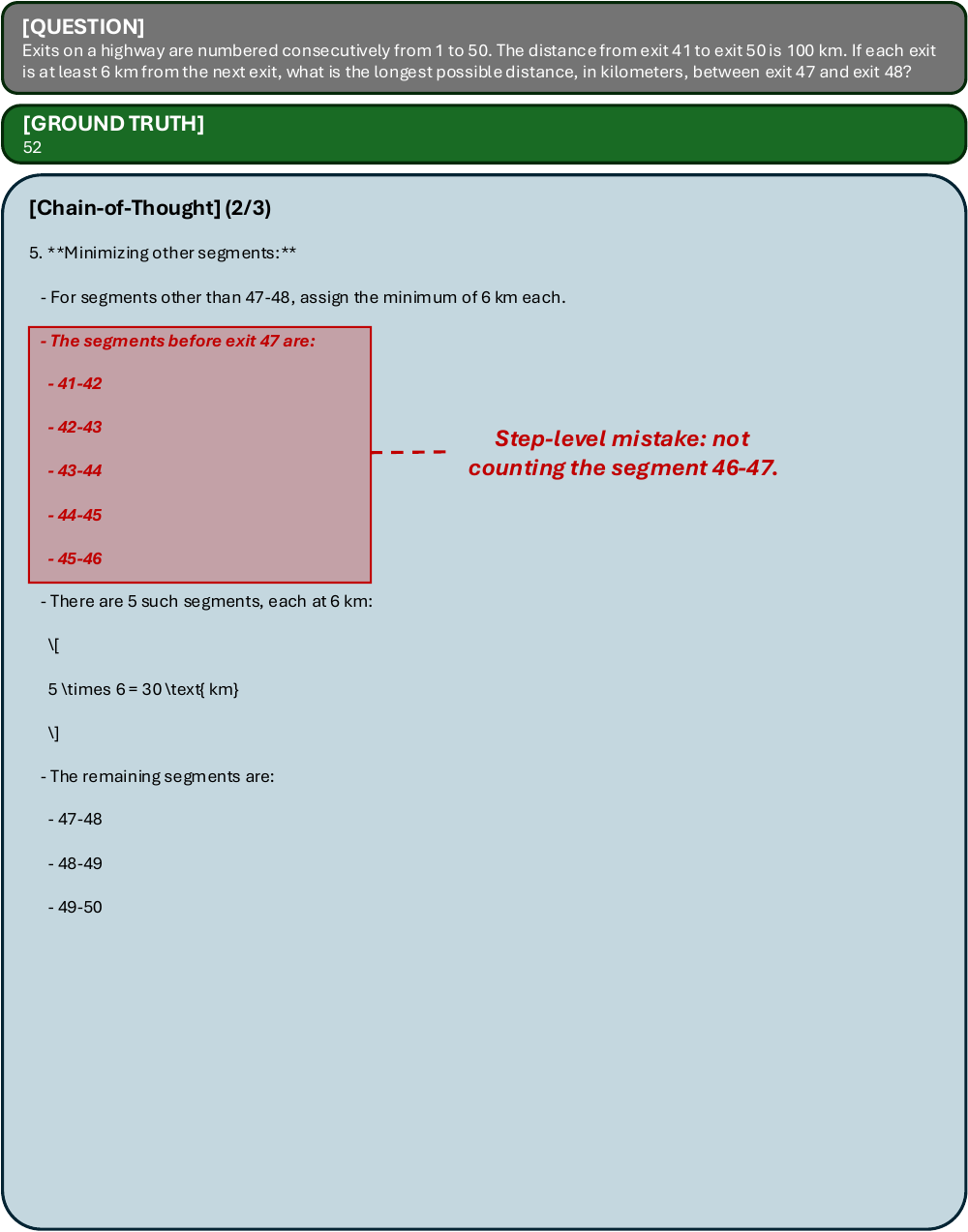}
\caption{
    \textbf{Qualitative Result, Concrete Example (2/7):} Chain-of-Thought Response (2/3).
}
\vspace{-0.0em}
\label{fig:qual-2}
\end{figure}

\begin{figure}[t] 
\centering 
\vspace{0em}
\includegraphics[width=1\textwidth]{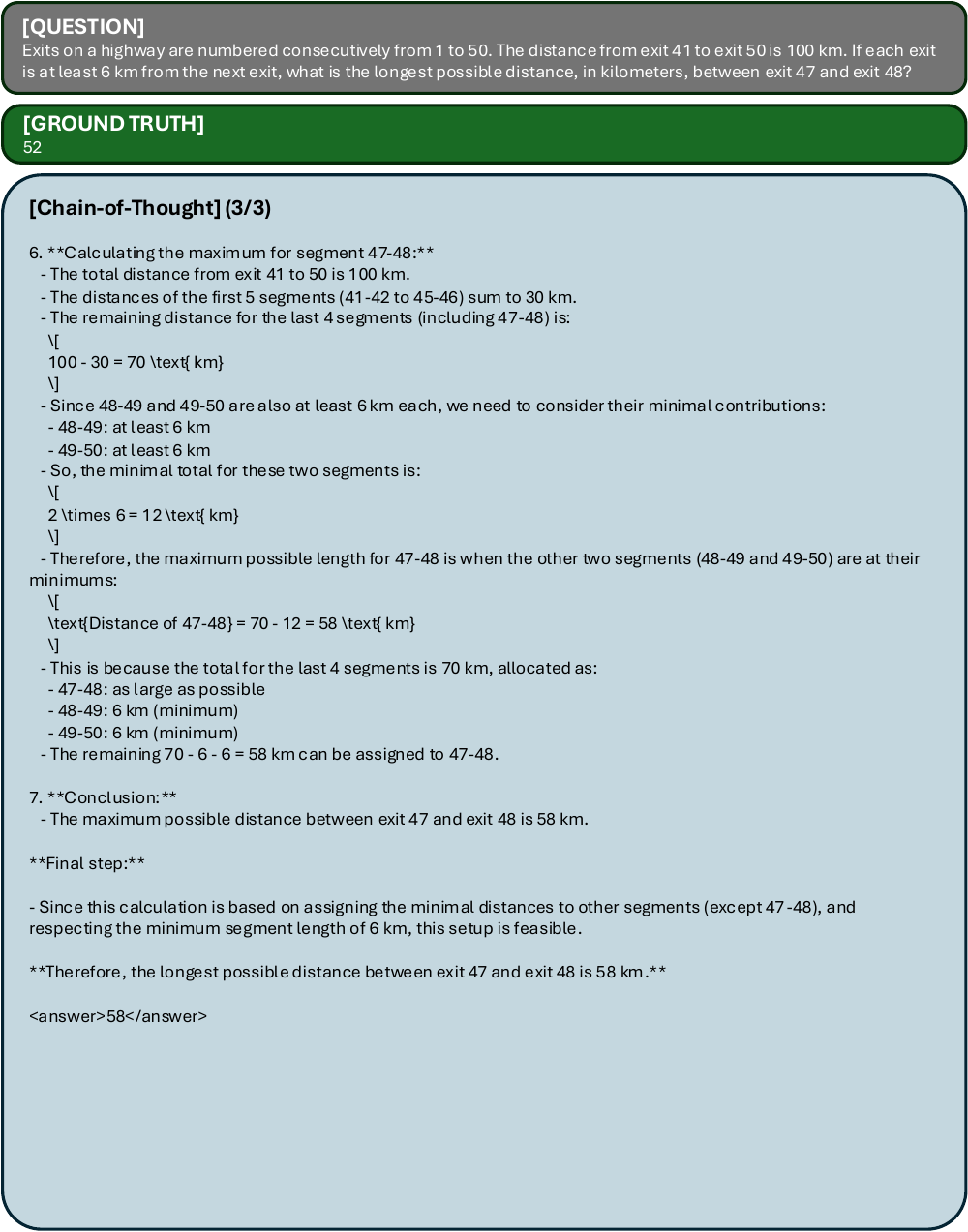}
\caption{
    \textbf{Qualitative Result, Concrete Example (3/7):} Chain-of-Thought Response (3/3).
}
\vspace{-0.0em}
\label{fig:qual-3}
\end{figure}

\begin{figure}[t] 
\centering 
\vspace{0em}
\includegraphics[width=1\textwidth]{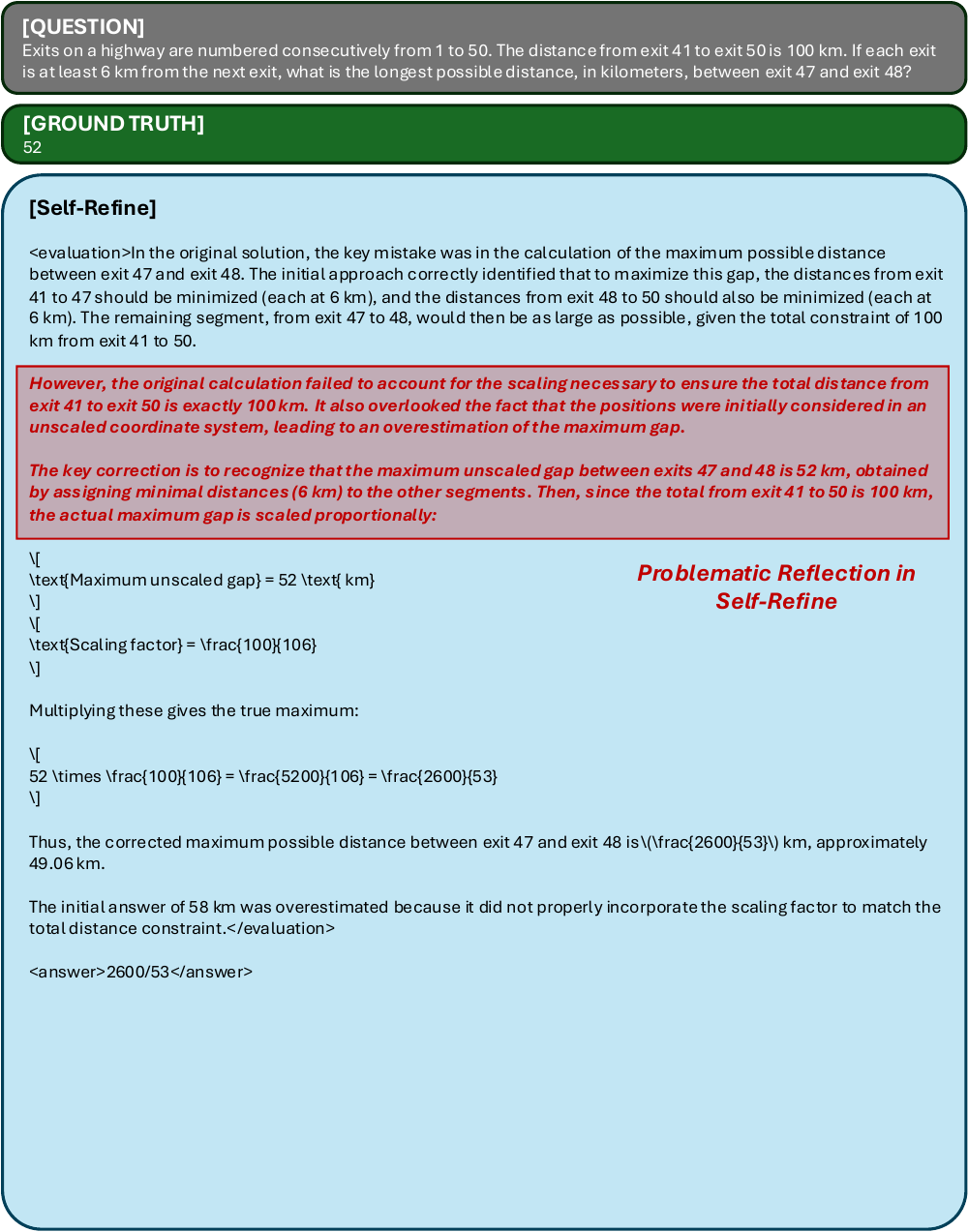}
\caption{
    \textbf{Qualitative Result, Concrete Example (4/7):} Self-Refine.
}
\vspace{-0.0em}
\label{fig:qual-4}
\end{figure}

\begin{figure}[t] 
\centering 
\vspace{0em}
\includegraphics[width=1\textwidth]{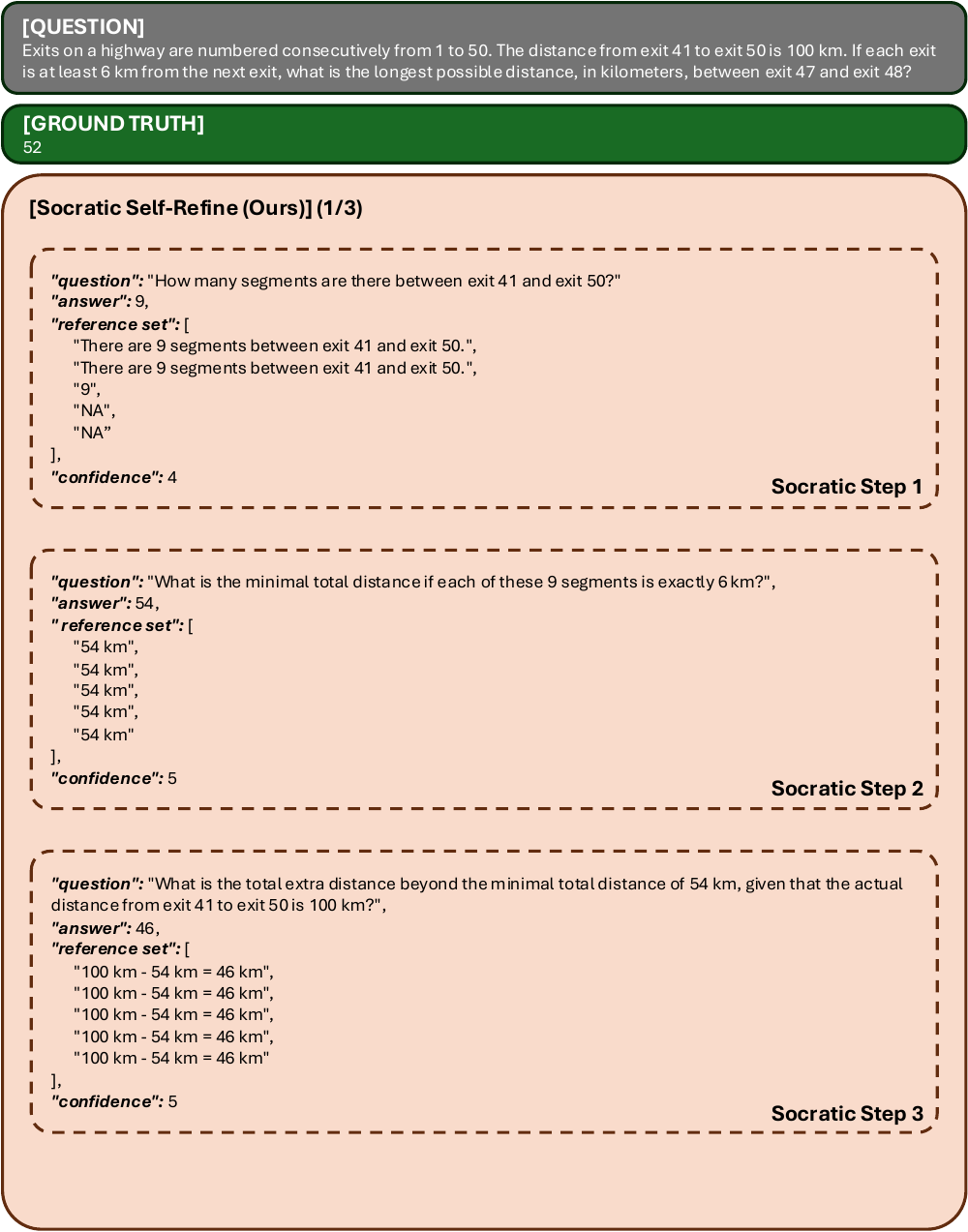}
\caption{
    \textbf{Qualitative Result, Concrete Example (5/7):} Socratic Self-Refine (\ours, Ours) (1/3).
}
\vspace{-0.0em}
\label{fig:qual-5}
\end{figure}

\begin{figure}[t] 
\centering 
\vspace{0em}
\includegraphics[width=1\textwidth]{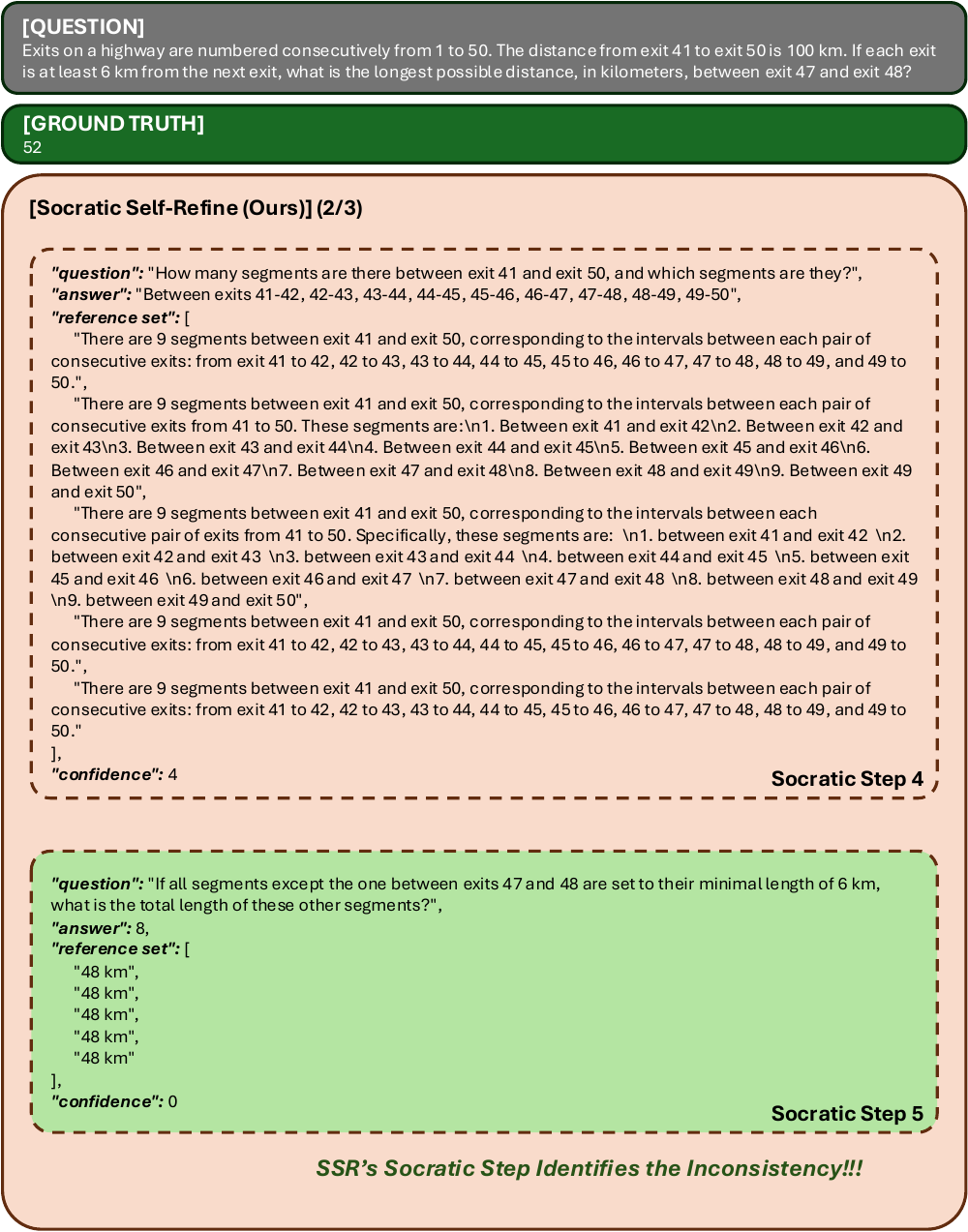}
\caption{
    \textbf{Qualitative Result, Concrete Example (6/7):} Socratic Self-Refine (\ours, Ours) (2/3).
}
\vspace{-0.0em}
\label{fig:qual-6}
\end{figure}

\begin{figure}[t] 
\centering 
\vspace{0em}
\includegraphics[width=1\textwidth]{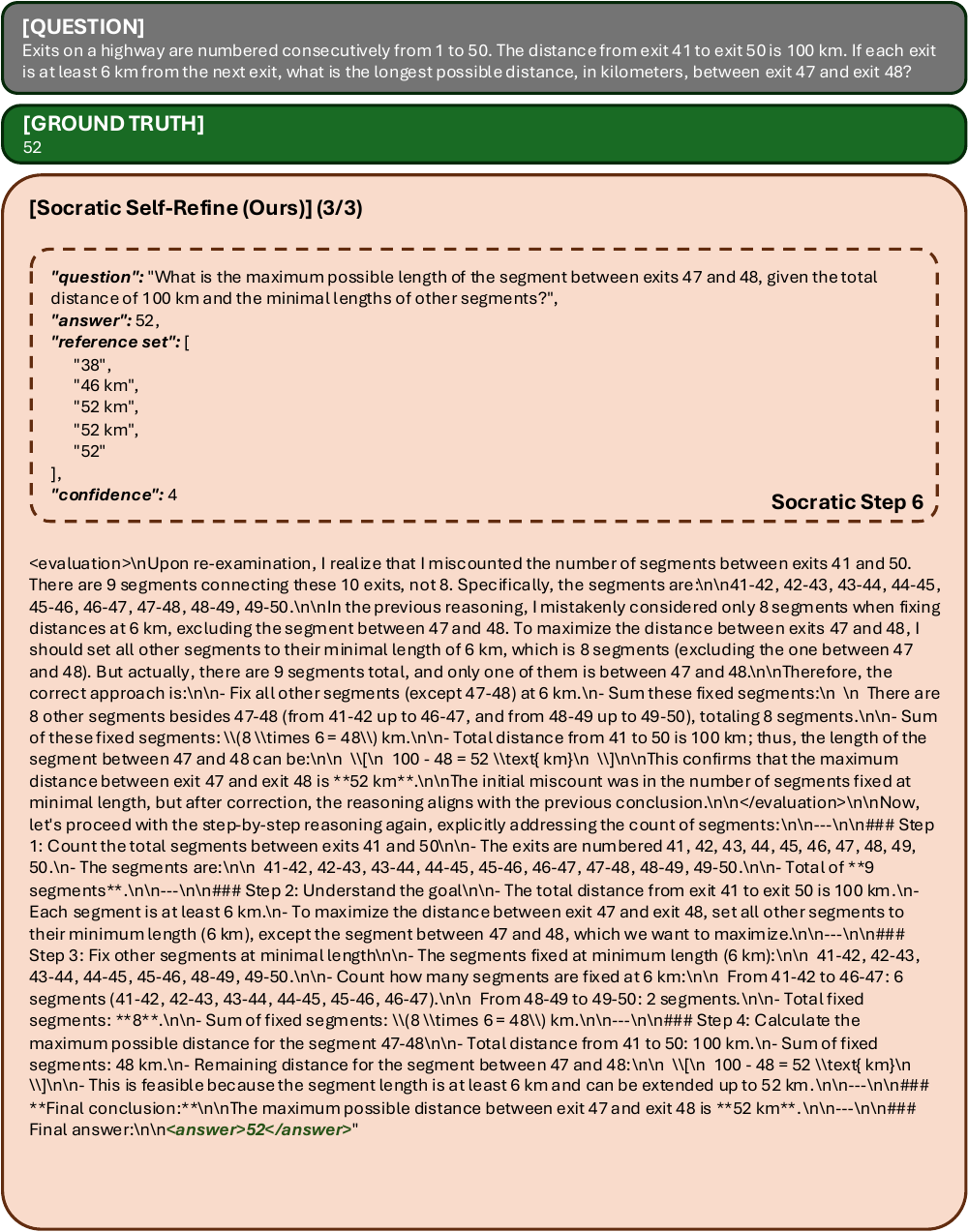}
\caption{
    \textbf{Qualitative Result, Concrete Example (7/7):} Socratic Self-Refine (\ours, Ours) (3/3).
}
\vspace{-0.0em}
\label{fig:qual-7}
\end{figure}

In this section, we present a concrete example to illustrate how \ours performs step-level verification and refinement. The example is taken from a MATH-Level-5 question, with \gptnano serving as the backbone LLM. {\textbf{The results are shown in \Figref{fig:qual-1}-\ref{fig:qual-7}.}}

The qualitative example illustrates how \ours intervenes within the reasoning process rather than only at the final output. By decomposing reasoning into Socratic steps, estimating step-level confidence, and selectively refining weak points, \ours prevents local errors from propagating through the reasoning trace. Compared to standard Self-Refine, which often produces brittle trajectories, \ours provides interpretable and controllable corrections that lead to more reliable outcomes. These case studies complement our quantitative results and highlight the practical value of \ours as a step-level refinement framework.

\end{document}